\documentclass[10pt,twocolumn,letterpaper]{article}
\usepackage{titling}

\usepackage[pagenumbers]{cvpr} % To force page numbers, e.g. for an arXiv version

\usepackage{graphicx}
\usepackage{amsmath}
\usepackage{amssymb}
\usepackage{booktabs}
\usepackage{multirow}

\usepackage[pagebackref,breaklinks,colorlinks]{hyperref}

\usepackage{color, colortbl}
\definecolor{LightGray}{rgb}{0.92,0.92,0.92}
\definecolor{Gray1}{rgb}{0.95,0.95,0.95}
\definecolor{Gray2}{rgb}{0.9,0.9,0.9}
\definecolor{darkblue}{rgb}{0,0.592,0.655}%{0, 151, 167}
\definecolor{darkyellow}{rgb}{0.655,0.608,0}%{167, 155, 0}
\usepackage{times,graphicx,tabulary,multirow,xspace,xcolor}

\usepackage{pifont}
\newcommand{\cmark}{\color{gray}\ding{51}}%
\newcommand{\xmark}{\color{gray}\ding{55}}%
\newcommand{\modelname}{ReCo}

\makeatletter
\DeclareRobustCommand\onedot{\futurelet\@let@token\@onedot}
\def\@onedot{\ifx\@let@token.\else.\null\fi\xspace}

\def\eg{\emph{e.g}\onedot} 
\def\ie{\emph{i.e}\onedot} 
 
\def\etc{\emph{etc}\onedot} \def\vs{\emph{vs}\onedot}

\usepackage[capitalize]{cleveref}
\crefname{section}{Sec.}{Secs.}
\Crefname{section}{Section}{Sections}
\Crefname{table}{Table}{Tables}
\crefname{table}{Tab.}{Tabs.}

\usepackage{wrapfig}

\definecolor{LightCyan}{rgb}{0.9059,0.9961,1}
\usepackage{multirow}
\usepackage{graphicx}
\usepackage[font=small,labelfont=bf,aboveskip=1pt,belowskip=-13pt]{caption}  % set global caption size
\usepackage{makecell}
\usepackage{tabulary}
\definecolor{demphcolor}{RGB}{144,144,144}

\usepackage{pifont}% http://ctan.org/pkg/pifont for xmark and cmark
\newlength\savewidth

\newcommand{\tablestyle}[2]{\setlength{\tabcolsep}{#1}\renewcommand{\arraystretch}{#2}\centering\small}
\makeatletter\renewcommand\paragraph{\@startsection{paragraph}{4}{\z@}
  {.5em \@plus1ex \@minus.2ex}{-.5em}{\normalfont\normalsize\bfseries}}\makeatother
\newdimen\abovecrulesep
\newdimen\belowcrulesep
\makeatletter
\patchcmd{\@@@cmidrule}{\aboverulesep}{\abovecrulesep}{}{}
\patchcmd{\@xcmidrule}{\belowrulesep}{\belowcrulesep}{}{}
\makeatother
\def\cvprPaperID{****} % *** Enter the CVPR Paper ID here
\def\confName{****}
\def\confYear{****}

\begin{document}

\title{ReCo: Region-Controlled Text-to-Image Generation}

\author{Zhengyuan Yang, Jianfeng Wang, Zhe Gan, Linjie Li, Kevin Lin, \\ Chenfei Wu, Nan Duan, Zicheng Liu, Ce Liu, Michael Zeng, Lijuan Wang \\
Microsoft \\ 
{\tt\footnotesize \{zhengyang,jianfw,zhe.gan,lindsey.li,keli,chewu,nanduan,zliu,ce.liu,nzeng,lijuanw\}@microsoft.com}
}

% \maketitle
\twocolumn[{\renewcommand\twocolumn[1][]{#1}\maketitle
\centering
\small
\vspace{-2em}
\includegraphics[width=0.95\textwidth]{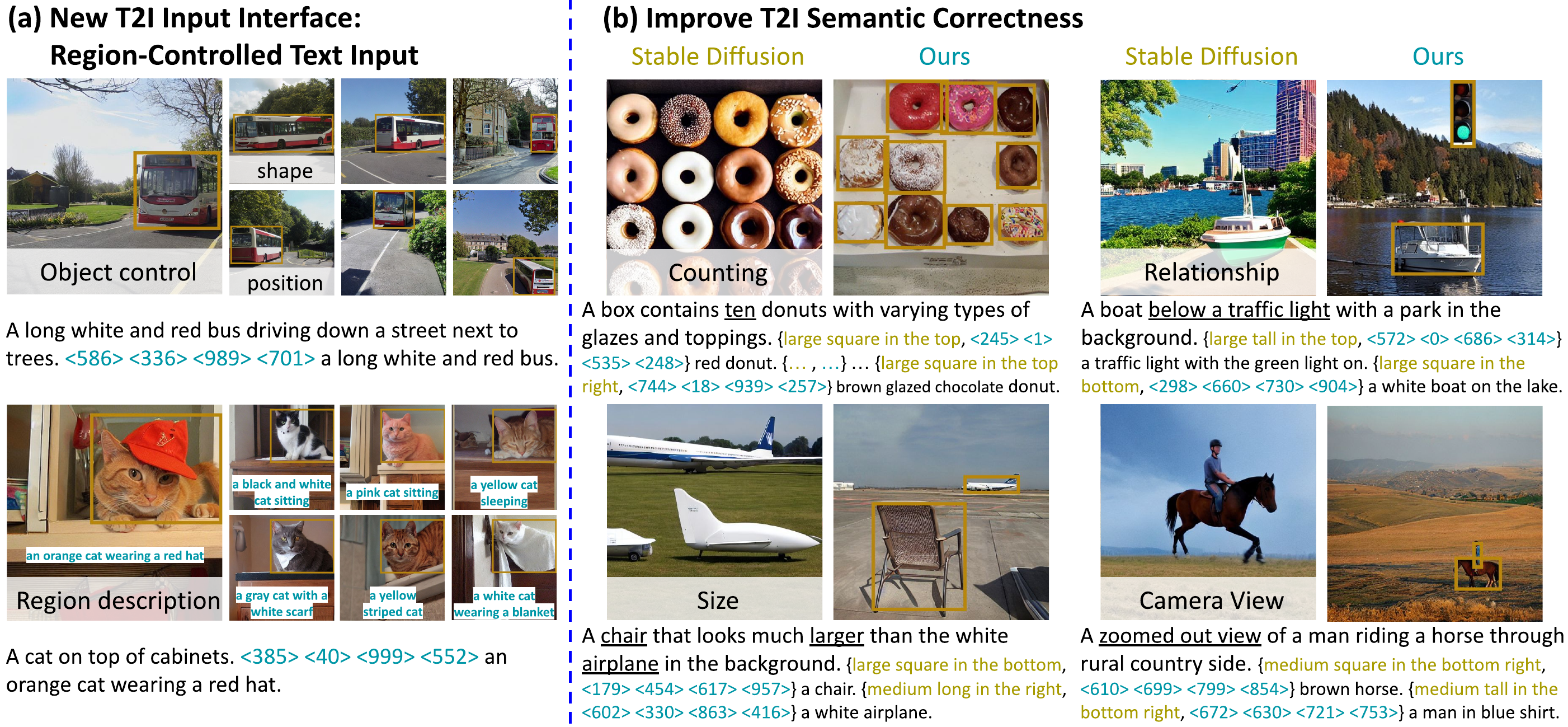} \\ % 0.80 is 14cm; w=17.5cm
\vspace{0.5em}
\captionof{figure}{
\textbf{(a)} \emph{\modelname} extends pre-trained text-to-image models (Stable Diffusion~\cite{rombach2022high}) with an extra set of input position tokens ({\color{darkblue} in dark blue color}) that represent quantized spatial coordinates. Combining position and text tokens yields the region-controlled text input, which can specify an open-ended regional description precisely for any image region.
\textbf{(b)} With the region-controlled text input, \emph{\modelname} can better control the object count/relationship/size properties and improve the T2I semantic correctness. We empirically observe that {\color{darkblue}position tokens} are less likely to get overlooked than {\color{darkyellow}positional text words}, especially when the input query is complicated or describes an unusual scene.
}
\label{fig:teaser}
\vspace{3em}}]

%%%%%%%%% ABSTRACT
\begin{abstract}
\vspace{-3mm}

Recently, large-scale text-to-image (T2I) models have shown impressive performance in generating high-fidelity images, but with limited controllability, e.g., precisely specifying the content in a specific region with a free-form text description.
In this paper, we propose an effective technique for such regional control in T2I generation. 
We augment T2I models' inputs with an extra set of position tokens, which represent the quantized spatial coordinates. Each region is specified by four position tokens to represent the top-left and bottom-right corners, followed by an open-ended natural language regional description. Then, we fine-tune a pre-trained T2I model with such new input interface.
Our model, dubbed as \modelname~(Region-Controlled T2I), enables the region control for arbitrary objects described by open-ended regional texts rather than by object labels from a constrained category set.
Empirically, \modelname~achieves better image quality than the T2I model strengthened by positional words (FID: $8.82\rightarrow7.36$, SceneFID: $15.54\rightarrow6.51$ on COCO), together with objects being more accurately placed, amounting to a $20.40\%$ region classification accuracy improvement on COCO.
Furthermore, we demonstrate that \modelname~can better control the object count, spatial relationship, and region attributes such as color/size, with the free-form regional description. Human evaluation on PaintSkill shows that \modelname~is $+19.28\%$ and $+17.21\%$ more accurate in generating images with correct object count and spatial relationship than the T2I model.
\end{abstract}
\vspace{-4mm}
\section{Introduction}
%%%%%%%%%%%%%%%%%%%%%%%%%%%%%%%%%%%%%%%%

\begin{figure}[t]
\centering
\includegraphics[width=.48\textwidth]{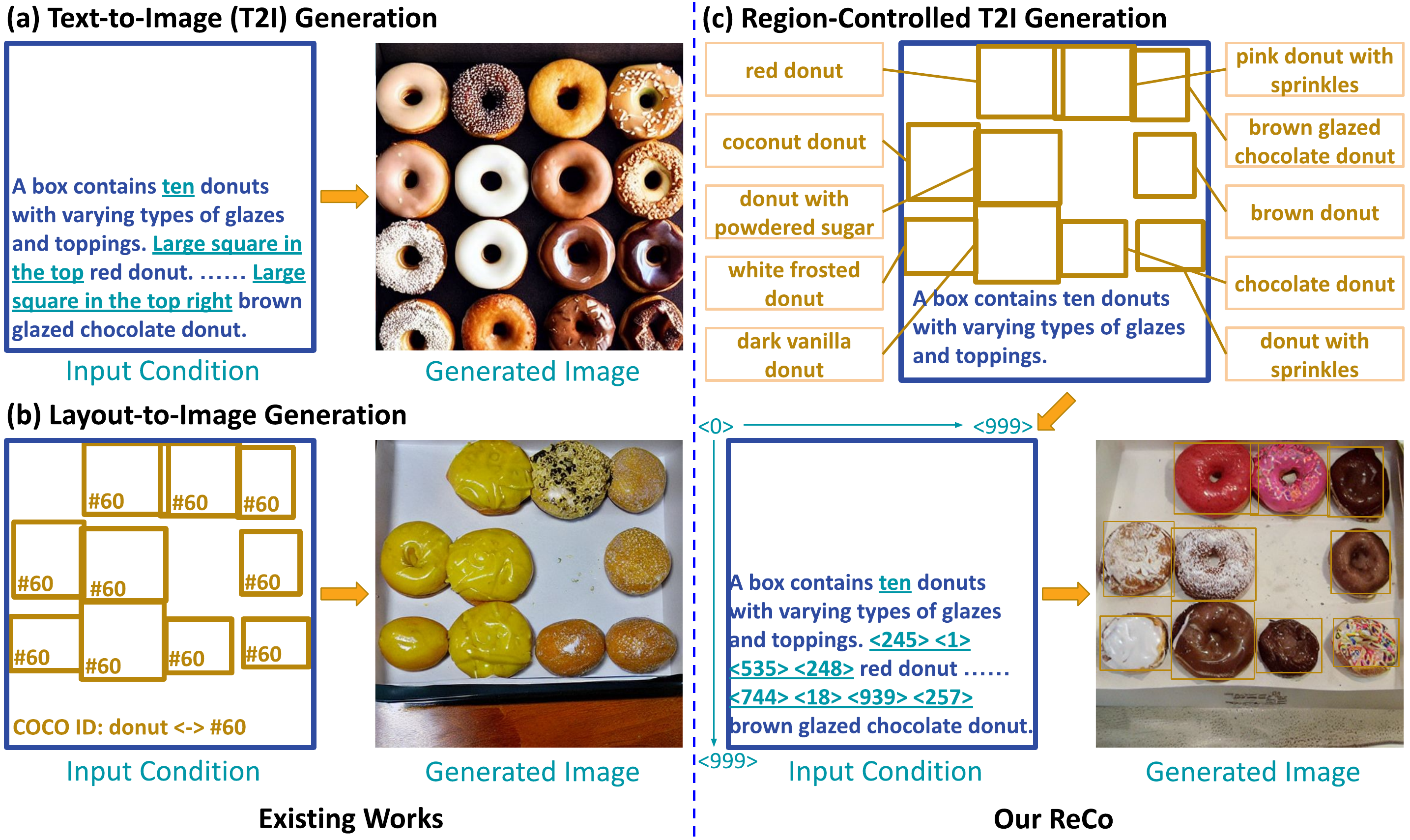}
% \vspace{-0.2in}
\caption[Caption for LOF]{
    \textbf{(a)} With positional words (\eg, bottom/top/left/right and large/small/tall/long), the T2I model (Stable Diffusion~\cite{rombach2022high}) does not manage to create objects with desired properties.
    \textbf{(b)} Layout-to-image generation~\cite{zhao2019image,sun2019image,li2020bachgan,rombach2022high} takes all object boxes and labels as the input condition, but only works well with constrained object labels.
    \textbf{(c)} Our \modelname~model synergetically combines the text and box referring, allowing users to specify an open-ended regional text description precisely at any image region.
	}
\label{fig:intro}
\end{figure}

Text-to-image (T2I) generation aims to generate faithful images based on an input text query that describes the image content. By scaling up the training data and model size, large T2I models~\cite{ramesh2022hierarchical,saharia2022photorealistic,yu2022scaling,rombach2022high} have recently shown remarkable capabilities in generating high-fidelity images. 
However, the text-only query allows limited controllability, \eg, precisely specifying the content in a specific region. % object positions or regional descriptions. 
The naive way of using position-related text words, such as ``top left'' and ``bottom right,'' often results in ambiguous and verbose input queries, as shown in Figure~\ref{fig:intro} (a).
Even worse, when the text query becomes long and complicated, or describes an unusual scene, T2I models~\cite{ramesh2022hierarchical,yu2022scaling} might overlook certain details and rather follow the visual or linguistic training prior. These two factors together make region control difficult. To get the desired image, users usually need to try a large number of paraphrased queries and pick an image that best fits the desired scene. The process known as ``prompt engineering'' is time-consuming and often fails to produce the desired image.

The desired region-controlled T2I generation is closely related to the layout-to-image generation~\cite{zhao2019image,sun2019image,li2020bachgan,li2021image,frolov2021attrlostgan,yang2022modeling,rombach2022high,fan2022frido}.
As shown in Figure~\ref{fig:intro} (b), layout-to-image models take all object bounding boxes with labels from a close set of object vocabulary~\cite{lin2014microsoft} as inputs. Despite showing promise in region control, they can hardly understand free-form text inputs, nor the region-level combination of open-ended text descriptions and spatial positions. 
The two input conditions of text and box provide complementary referring capabilities. Instead of separately modeling them as in text-to-image and layout-to-image generations, we study ``region-controlled T2I generation'' that seamlessly combines these two input conditions. As shown in Figure~\ref{fig:intro} (c), the new input interface allows users to provide open-ended descriptions for arbitrary image regions, such as precisely placing a ``brown glazed chocolate donut'' in a specific area.

To this end, we propose \modelname~(Region-Controlled T2I) that extends pre-trained T2I models to understand spatial coordinate inputs.
The core idea is to introduce an extra set of input position tokens 
to indicate the spatial positions. 
The image width/height is quantized uniformly into $N_\text{bins}$ bins. Then, any float-valued coordinate can be approximated and tokenized by the nearest bin. With an extra embedding matrix ($E_p$), the position token can be mapped onto the same space as the text token.
Instead of designing a text-only query with positional words ``in the \emph{top} red donut'' as in Figure~\ref{fig:intro} (a), \modelname~takes region-controlled text inputs ``$\tiny{<}x_\text{1}\tiny{>},\tiny{<}y_\text{1}\tiny{>}, \tiny{<}x_\text{2}\tiny{>},\tiny{<}y_\text{2}\tiny{>}$ red donut,'' where $\tiny{<}x\tiny{>}$,$\tiny{<}y\tiny{>}$ are the position tokens followed by the corresponding free-form text description. 
We then fine-tune a pre-trained T2I model with $E_p$ to generate the image from the extended input query. 
To best preserve the pre-trained T2I capability, \modelname~training is designed to be similar to the T2I pre-training, \ie, introducing minimal extra model parameters ($E_p$), jointly encoding position and text tokens with the text encoder, and prefixing the image description before the extended regional descriptions in the input query.

%%%%%%%%%%%%%%%%%%%%%%%%%%%%%%%%%%%%%%%%
%%%%%%%%%%%%%%%%%%%%%%%%%%%%%%%%%%%%%%%%%%%%%%%%%%%%%%%%%%%%%%%%%%
Figure~\ref{fig:teaser} visualizes \modelname's use cases and capabilities. As shown in Figure~\ref{fig:teaser} (a), we can easily control the view (front/side) and type (single-/double-deck) of the ``bus'' by tweaking position tokens in region-controlled text inputs.
Position tokens also allow the user to provide free-form regional descriptions, such as ``an orange cat wearing a red hat'' at a specific location.
Furthermore, we empirically observe that position tokens are less likely to get overlooked or misunderstood than text words. As shown in Figure~\ref{fig:teaser} (b), \modelname~has better control over object count, spatial relationship, and size properties, especially when the query is long and complicated, or describes a scene that is less common in real life. In contrast, T2I models~\cite{rombach2022high} may struggle with generating scenes with correct object counts (``ten''), relationships (``boat below traffic light''), relative sizes (``chair larger than airplane''), and camera views (``zoomed out'').

To evaluate the region control, we design a comprehensive experiment benchmark based on a pre-trained regional object classifier and an object detector. The object classifier is applied on the generated image regions, while the detector is applied on the whole image. A higher accuracy means a better alignment between the generated object layout and the region positions in user queries.
On the COCO dataset~\cite{lin2014microsoft}, \modelname~shows a better object classification accuracy ($42.02\%\rightarrow62.42\%$) and detector averaged precision ($2.3\rightarrow32.0$), compared with the T2I model with carefully designed positional words. For image generation quality, \modelname~improves the FID from $8.82$ to $7.36$, and SceneFID from $15.54$ to $6.51$.
Furthermore, human evaluations on PaintSkill~\cite{cho2022dall} show $+19.28\%$ and $+17.21\%$ accuracy gain in more correctly generating the query-described object count and spatial relationship, indicating \modelname's capability in helping T2I models to generate challenging scenes.

\vspace{1pt}
Our contributions are summarized as follows.
%%%%%%%%%%%%%%%%%%%%%%%%%%%%%%%%%%%%%%%%
% \vspace{-3pt}
\vspace{-4pt}
\begin{itemize}
\setlength\itemsep{-1.5pt}
\item We propose \modelname~that extends pre-trained T2I models to understand coordinate inputs. Thanks to the introduced position tokens in the region-controlled input query, users can easily specify free-form regional descriptions in arbitrary image regions.
\item We instantiate \modelname~based on Stable Diffusion. Extensive experiments show that \modelname~strictly follows the regional instructions from the input query, and also generates higher-fidelity images. 
\item We design a comprehensive evaluation benchmark to validate \modelname's region-controlled T2I generation capability. \modelname~significantly improves both the region control accuracy and the image generation quality over a wide range of datasets and designed prompts.
\end{itemize}
\section{Related Work}
%%%%%%%%%%%%%%%%%%%%%%%%%%%%%%%%%%%%%%%%
\noindent\textbf{Text-to-image generation.}
Text-to-image (T2I) generation aims to generate a high-fidelity image based on an open-ended image description. Early studies adopt conditional GANs~\cite{reed2016generative,zhang2017stackgan,zhang2018stackgan++,xu2018attngan,zhang2021cross} for T2I generation. Recent studies have made tremendous advances by scaling up both the data and model size, based on either auto-regressive~\cite{gafni2022make,yu2022scaling} or diffusion-based models~\cite{rombach2022high,ramesh2022hierarchical,saharia2022photorealistic}. 
We build our study on top of the successful large-scale pre-trained T2I models, and explore how to better control the T2I generation by extending a pre-trained T2I model to understand position tokens.

%%%%%%%%%%%%%%%%%%%%%%%%%%%%%%%%%%%%%%%%
\vspace{2mm}
\noindent\textbf{Layout-to-image generation.}
Layout-to-image studies aim to generate an image from a complete layout, \ie, all bounding boxes and the paired object labels. Early studies~\cite{zhao2019image,sun2019image,li2020bachgan,li2021image,frolov2021attrlostgan} adopt GAN-based approaches by properly injecting the encoded layout as the input condition. Recent studies successfully apply the layout query as the input condition to the auto-regressive framework~\cite{gafni2022make,yang2022modeling} and diffusion models~\cite{rombach2022high,fan2022frido}. Our study is related to the layout-to-image generation as both directions require the model to understand coordinate inputs. The major difference is that our design synergetically combines text and box to help T2I generation. Therefore, \modelname~can take open-ended regional descriptions and benefit from large-scale T2I pre-training.

% %%%%%%%%%%%%%%%%%%%%%%%%%%%%%%%%%%%%%%%%
\vspace{2mm}
\noindent\textbf{Unifying open-ended text and localization conditions.}
Previous studies have explored unifying open-ended text descriptions with localization referring (box, mask, mouse trace) as the input generation condition. One modeling approach~\cite{reed2016generative,li2020image,pavllo2020controlling,hinz2019generating,gafni2022make,huang2022multimodal} is to separately encode the image description in T2I and the layout condition in layout-to-image, and trains a model to jointly condition on both input types. TRECS~\cite{koh2021text} takes mouse traces in the localized narratives dataset~\cite{pont2020connecting} to better ground open-ended text descriptions with a localized position. Other than taking layout as user-generated inputs, previous studies~\cite{hong2018inferring,li2019object} have also explored predicting layout from text to ease the T2I generation of complex scenes. Unlike the motivation of training another conditional generation model parallel to T2I and layout-to-image, we explore how to effectively extend pre-trained T2I models to understand region queries, leading to significantly better controllability and generation quality than training from scratch. In short, we position \modelname~as an improvement for T2I by providing a more flexible input interface and alleviating controllability issues, \eg, being difficult to override data prior when generating unusual scenes, and overlooking words in complex queries.

%%%%%%%%%%%%%%%%%%%%%%%%%%%%%%%%%%%%%%%%
\begin{figure*}[t]
\centering
\includegraphics[width=.95\textwidth]{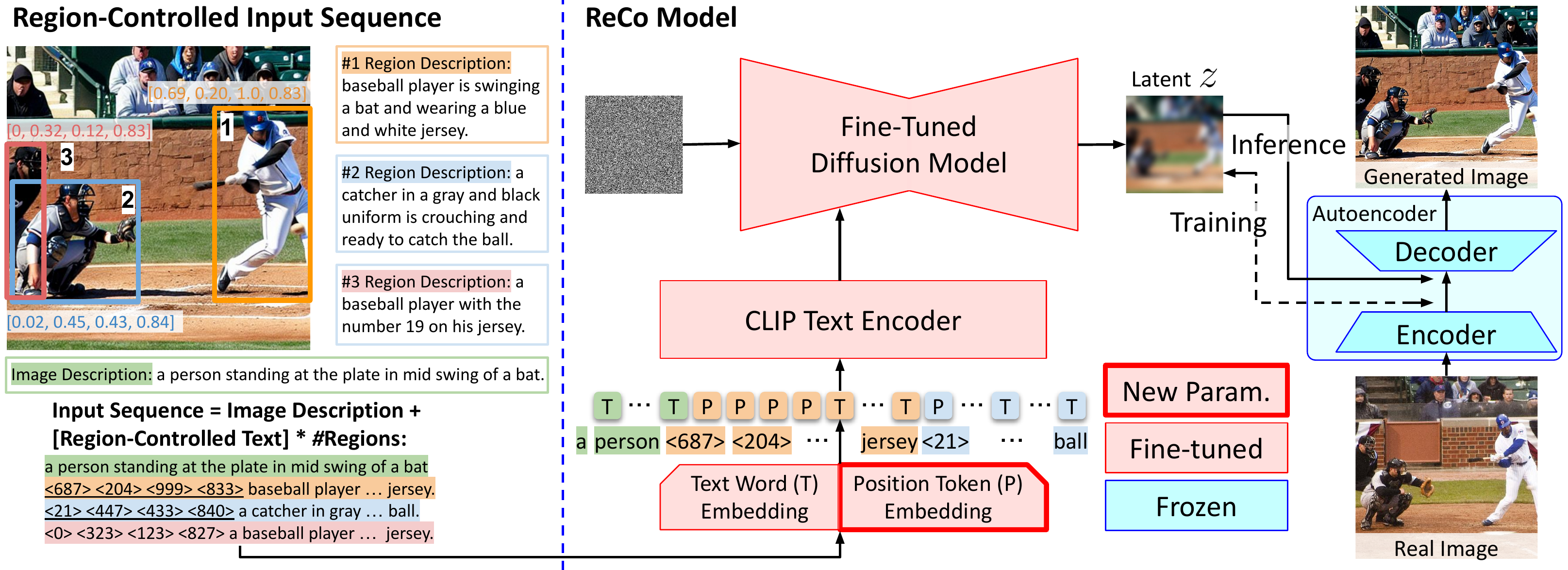}
    \caption{
    \modelname~extends Stable Diffusion~\cite{rombach2022high} with position tokens $P$ to support open-ended text description at both image- and region-level. We minimize the amount of introduced new model parameters (\ie, position token embedding $E_p$) to best preserve the pre-trained T2I capability. The diffusion model and text encoder are fine-tuned together to support the extended position token inputs.
	}
\label{fig:arch}
\end{figure*}
%%%%%%%%%%%%%%%%%%%%%%%%%%%%%%%%%%%%%%%%%%%%%%%%%%%%%%%%%%%%%%%%%%
\section{\modelname~Model}
Region-Controlled T2I Generation (\modelname) extends T2I models with the ability to understand coordinate inputs. 
The core idea is to design a unified input token vocabulary containing both text words and position tokens to allow accurate and open-ended regional control. 
By seamlessly mixing text and position tokens in the input query, \modelname~obtains the best from the two worlds of text-to-image and layout-to-image, \ie, the abilities of free-form description and precise position control.
In this section, we present our \modelname~implementation based on the open-sourced Stable Diffusion (SD)~\cite{rombach2022high}. We start with the SD preliminaries in Section~\ref{sec:sd} and introduce the core \modelname~design in Section~\ref{sec:poco}.

%%%%%%%%%%%%%%%%%%%%%%%%%%%%%%%%%%%%%%%%
\subsection{Preliminaries}
\label{sec:sd}
We take Stable Diffusion as an example to introduce the T2I model that \modelname~is built upon. Stable Diffusion is developed upon the Latent Diffusion Model~\cite{rombach2022high}, and consists of an auto-encoder, a U-Net~\cite{ronneberger2015u} for noise estimation, and a CLIP ViT-L/14 text encoder. For the auto-encoder, the encoder $\mathcal{E}$ with a down-sampling factor of $8$ encodes the image $x$ into a latent representation $z=\mathcal{E}(x)$ that the diffusion process operates on, and the decoder $\mathcal{D}$ reconstructs the image $\hat{x}=\mathcal{D}(z)$ from the latent $z$. 
U-Net~\cite{ronneberger2015u} is conditioned on denoising timestep $t$ and text condition $\tau_{\theta}(y(T))$, where $y(T)$ is the input text query with text tokens $T$ and $\tau_{\theta}$ is the CLIP ViT-L/14 text encoder~\cite{radford2021learning} that projects a sequence of tokenized texts into the sequence embedding.

The core motivation of \modelname~is to explore more effective and interaction-friendly conditioning signals $y$, while best preserving the pre-trained T2I capability. Specifically, \modelname~extends text tokens with an extra vocabulary specialized for spatial coordinate referring, \ie, position tokens $P$, which can be seamlessly used together with text words $T$ in a single input query $y$. \modelname~aims to show the benefit of synergetically combining text and position conditions for region-controlled T2I generation. % with the proposed approach.

%%%%%%%%%%%%%%%%%%%%%%%%%%%%%%%%%%%%%%%%
\subsection{Region-Controlled T2I Generation}
\label{sec:poco}
\noindent\textbf{\modelname~input sequence.}
The text input in T2I generation provides a natural way of specifying the generation condition. However, text words may be ambiguous and verbose in providing regional specifications.
For a better input query, \modelname~introduces position tokens that can directly refer to a spatial position. Specifically, the position and size of each region can be represented by four floating numbers, \ie, top-left and bottom-right coordinates. By quantizing coordinates~\cite{chen2021pix2seq,yang2022unitab,wang2022ofa}, we can represent the region by four discrete position tokens $P$, $\tiny{<}x_\text{1}\tiny{>},\tiny{<}y_\text{1}\tiny{>}, \tiny{<}x_\text{2}\tiny{>},\tiny{<}y_\text{2}\tiny{>}$, arranged as a sequence similar to a short natural language sentence.
The left side of Figure~\ref{fig:arch} illustrates the \modelname~input sequence design. 
Same as T2I, we start the input query with the image description to make the best use of large-scale T2I pre-training.
The image description is followed by multiple region-controlled texts, \ie, the four position tokens and the corresponding open-ended regional description. The number of regional specifications is unlimited, allowing users to easily create complex scenes with more regions, or save time on composing input queries with fewer or even no regions.
\modelname~introduces position token embedding $E_p\in \mathbb{R}^{N_\text{bins}\times D}$ alongside the pre-trained text word embedding, where $N_\text{bins}$ is the number of the position tokens, and $D$ is the token embedding dimension.
The entire sequence is then processed jointly, and each token, either text or spatial, is projected into a $D$-dim token embedding. The pre-trained CLIP text encoder from Stable Diffusion takes the token embeddings in, and projects them as the sequence embedding that the diffusion model conditions on.

\vspace{2mm}
\noindent\textbf{ReCo fine-tuning.}
ReCo extends the text-only query $y(T)$ with text tokens $T$ into \modelname~input query $y(P,T)$ that combines the text word $T$ and position token $P$. We fine-tune the Stable Diffusion with the same latent diffusion modeling objective~\cite{rombach2022high}, following the notations in Section~\ref{sec:sd}:
\begin{equation*}
    L = \mathbb{E}_{\mathcal{E}(x),y(P,T),\epsilon\sim\mathcal{N}(0,1),t}\left[\lVert \epsilon-\epsilon_{\theta}(z_t, t, \tau_{\theta}(y(P,T))) \rVert_{2}^{2} \right],
\end{equation*}
where $\epsilon_{\theta}$ and $\tau_{\theta}$ are the fine-tuned network modules. All model parameters except position token embedding $E_p$ are initiated from the pre-trained Stable Diffusion model.
Both the image description and several regional descriptions are required for \modelname~model fine-tuning. For the training data, we run a state-of-the-part captioning model~\cite{wang2022git} on the cropped image regions (following the annotated bounding boxes) to get the regional descriptions. During fine-tuning, we resize the image with the short edge to $512$ and randomly crop a square region as the input image $x$. 
We will release the generated data and fine-tuned model for reproduction.

We empirically observe that \modelname~can well understand the introduced position tokens and precisely place objects at arbitrary specified regions. 
Furthermore, we find that position tokens can also help \modelname~better model long input sequences that contain multiple detailed attribute descriptions, leading to fewer detailed descriptions being neglected or incorrectly generated than the text-only query.
By introducing position tokens with a minimal change to the pre-trained T2I model, \modelname~obtains the desired region controllability while best preserving the appealing T2I capability.

%%%%%%%%%%%%%%%%%%%%%%%%%%%%%%%%%%%%%%%%
\begin{table*}[t]
\tablestyle{7pt}{1.0} 
\centering
{
\begin{tabular}{ l | c c c | c c c | c c }
    \hline
    \multirow{2}{*}{Method} &
    Image & Region & Region & \multicolumn{3}{c|}{Region Control Metrics ($\uparrow$)} & \multicolumn{2}{c}{Image Quality Metrics}\\
     & Descr. & Descr. & Position & AP & AP$_\text{50}$ & Object Acc. & SceneFID ($\downarrow$) & FID ($\downarrow$) \\
    \hline
    Real Images & - & - & - & \textcolor{gray}{36.8} & \textcolor{gray}{56.1} & \textcolor{gray}{71.41} & \textcolor{gray}{-} & \textcolor{gray}{-} \\
    \hline
     & \cmark & - & - & 0.7 & 2.0 & 26.75 & 35.80 & 13.40 \\
    SD V1.4 Zero-shot & \cmark & \cmark & - & 0.7 & 2.0 & 27.64 & 34.72 & 13.88 \\
     & \cmark & \cmark & Text & 0.6 & 1.8 & 28.15 & 32.86 & 14.57 \\
    \hline
    SD COCO Fine-tune: &  &  &  &  &  &  &  &   \\
    \modelname$_\text{Image Descr.}$ & \cmark & - & - & 0.9 & 2.6 & 29.12 & 27.78 & 10.44 \\
    \modelname$_\text{Region Descr.}$ & \cmark & \cmark & - & 1.0 & 2.9 & 32.32 & 24.88 & 9.11 \\
    \modelname$_\text{Position Word}$ & \cmark & \cmark & Text & 2.3 & 7.5 & 42.02 & 15.54 & 8.82 \\
    \modelname & \cmark & \cmark & \cmark & \textbf{32.0} & \textbf{52.4} & \textbf{62.42} & \textbf{6.51} & \textbf{7.36} \\
    \hline
\end{tabular}
}
\caption{Region control accuracy and image generation quality evaluations on the COCO (2014) 30k validation subset~\cite{lin2014microsoft,ramesh2021zero,yu2022scaling,wang2022ofa}.}
\label{table:coco}
\end{table*}
%%%%%%%%%%%%%%%%%%%%%%%%%%%%%%%%%%%%%%%%
%%%%%%%%%%%%%%%%%%%%%%%%%%%%%%%%%%%%%%%%
\begin{figure*}[t]
\centering
\includegraphics[width=.95\textwidth]{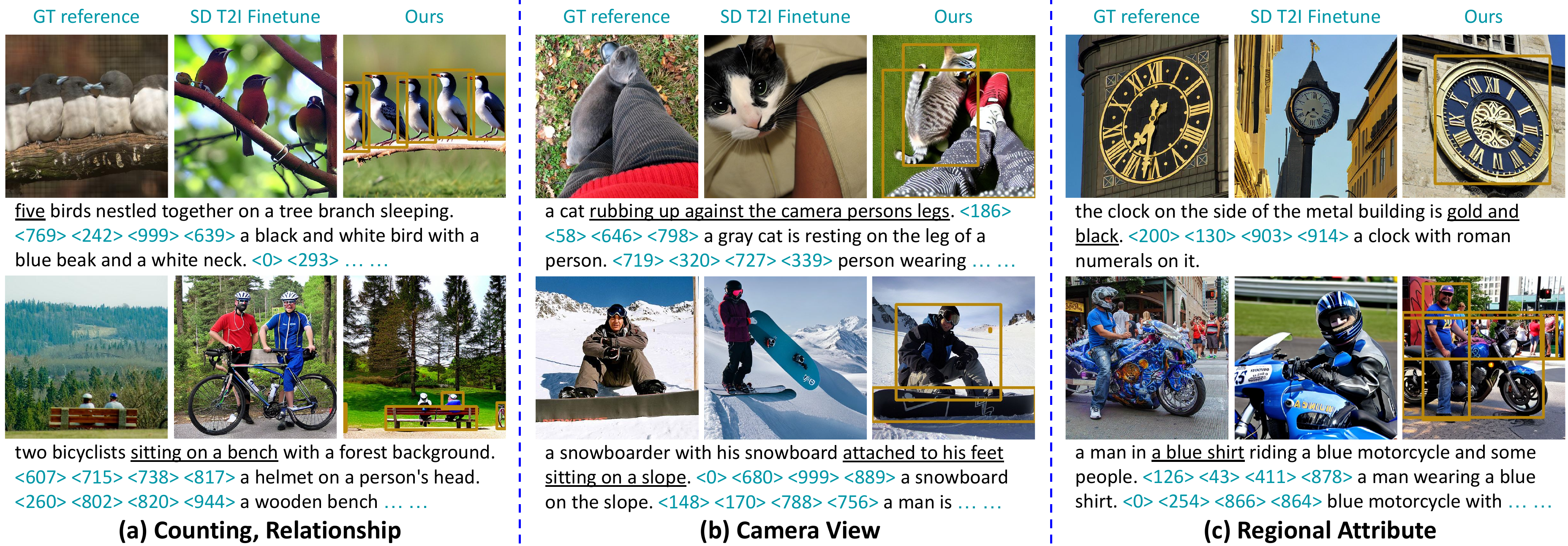}
    \caption{
    Qualitative results on COCO~\cite{lin2014microsoft}. \modelname's extra regional control (shown in the {\color{darkblue}dark blue color}) can improve T2I generation on (a) object counting and relationships, (b) images with unique camera views, and (c) images with detailed regional attribute descriptions. 
	}
\label{fig:visucoco}
\end{figure*}
%%%%%%%%%%%%%%%%%%%%%%%%%%%%%%%%%%%%%%%%%%%%%%%%%%%%%%%%%%%%%%%%%%
%%%%%%%%%%%%%%%%%%%%%%%%%%%%%%%%%%%%%%%%
\begin{figure*}[t]
\centering
\includegraphics[width=.95\textwidth]{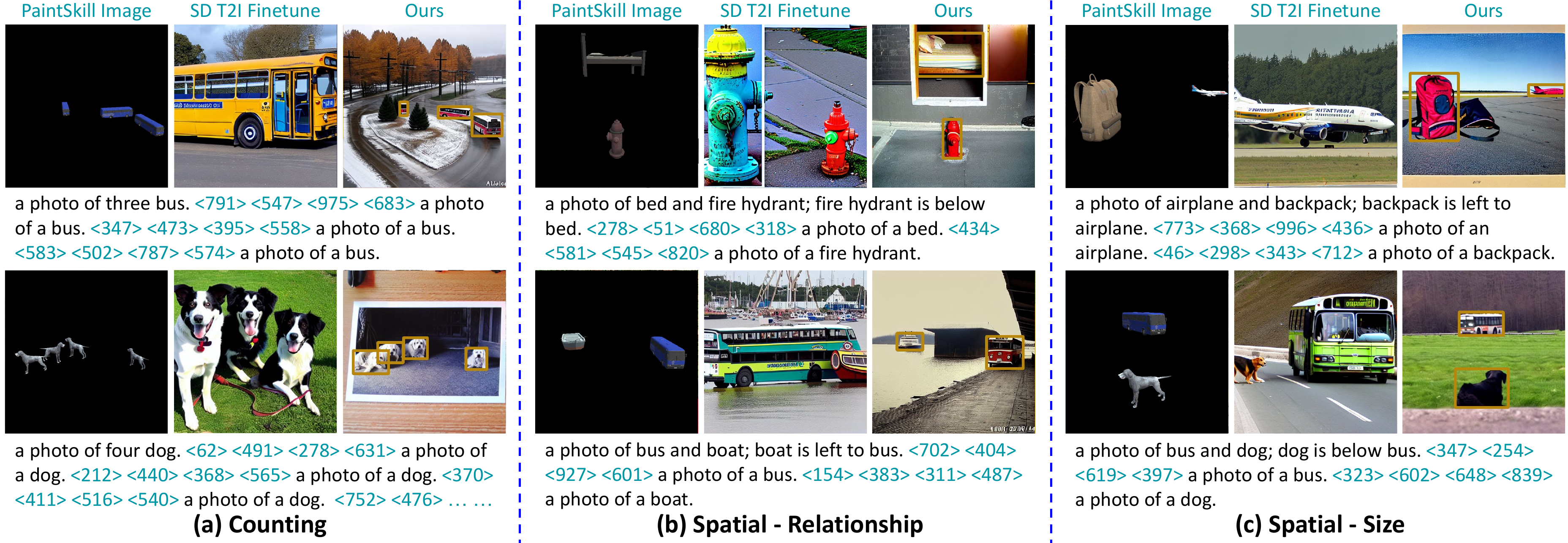}
    \caption{
    Qualitative results on PaintSkill~\cite{cho2022dall}. \modelname's extra regional control (shown in the {\color{darkblue}dark blue color}) can help T2I models more reliably generate scenes with exact object counts and unusual object relationships/relative sizes.
	}
\label{fig:visupaint}
\vspace{-0.03in}
\end{figure*}
%%%%%%%%%%%%%%%%%%%%%%%%%%%%%%%%%%%%%%%%%%%%%%%%%%%%%%%%%%%%%%%%%%
%%%%%%%%%%%%%%%%%%%%%%%%%%%%%%%%%%%%%%%%
\section{Experiments}
\subsection{Experiment Settings}
\noindent\textbf{Datasets.}
We evaluate the model on the COCO~\cite{lin2014microsoft,chen2015microsoft}, PaintSkill~\cite{cho2022dall}, and LVIS~\cite{gupta2019lvis} datasets. For input queries, we take image descriptions and boxes from the datasets~\cite{lin2014microsoft,gupta2019lvis,cho2022dall}, and generate regional descriptions with the same captioning model~\cite{wang2022git} on cropped regions. For COCO~\cite{lin2014microsoft,chen2015microsoft}, we follow the established setting in the T2I generation~\cite{ramesh2021zero,yu2022scaling,wang2022ofa} that reports the results on a subset of 30,000 captions sampled from the COCO 2014 validation set. We fine-tune stable diffusion with image-text pairs from the COCO 2014 training set. 
PaintSkill~\cite{cho2022dall} evaluates models' capabilities on following arbitrarily positioned boxes and generating images with the correct object type/count/relationship. We conduct the T2I inference with validation set prompts, which contain 1,050/2,520/3,528 queries for object recognition, counting, and spatial relationship skills, respectively.
LVIS~\cite{gupta2019lvis} tests if the model understands open-vocabulary regional descriptions, with the object categories unseen in the COCO fine-tuning data.
We report the results on the 4,809 LVIS validation images~\cite{gupta2019lvis} from the COCO 2017 validation set~\cite{johnson2018image,zhao2019image}. We do not further fine-tune the model when experimenting on PaintSkill and LVIS
to test the generalization capability in out-of-domain data. 

\noindent\textbf{Evaluation metrics.}
We evaluate \modelname~with metrics focused on region control accuracy and image generation quality. For region control accuracy, we use Object Classification Accuracy~\cite{zhao2019image} and DETR detector Average Precision (AP)~\cite{carion2020end}. Object accuracy trains a classifier with ground-truth (GT) image crops to classify the cropped regions on generated images. DETR detector AP detects objects on generated images and compares the results with input object queries. Thus, higher accuracy and AP can indicate a better layout alignment. For image generation quality, we use the Fr\'echet Inception Distance (FID)~\cite{heusel2017gans} to evaluate the image quality. We take SceneFID~\cite{sylvain2021object} as an indicator for region-level visual quality, which computes FID on the regions cropped based on input object boxes. We compute FID and SceneFID with the Clean-FID repo~\cite{parmar2022aliased} against center-cropped COCO images. We further conduct human evaluations on PaintSkill, due to the lack of GT images and effective automatic evaluation metrics. 

\noindent\textbf{Implementation details.}
We fine-tune \modelname~from the Stable Diffusion v1.4 checkpoint. We introduce $N=1000$ position tokens and increase the max length of the text encoder to 616. 
The batch size is 2048. We use AdamW optimizer~\cite{loshchilov2017decoupled} with a constant learning rate of $1e^{-4}$ to train the model for 20,000 steps, equivalent to around 100 epochs on COCO 2014 training set. The inference is conducted with 50 PLMS steps~\cite{liu2022pseudo}. We select a classifier-free guidance scale~\cite{ho2022classifier} that gives the best region control performance, \ie, 4.0 for \modelname~and 7.5 for original Stable Diffusion, detailed in Section~\ref{sec:ana}. We do not use CLIP image re-ranking.

%%%%%%%%%%%%%%%%%%%%%%%%%%%%%%%%%%%%%%%%
\subsection{Region-Controlled T2I Generation Results}
\label{sec:number}

\noindent\textbf{COCO.}
Table~\ref{table:coco} reports the region-controlled T2I generation results on COCO. The first row ``real images'' provides an oracle reference number on applicable metrics. The \emph{top part} of the table shows the results obtained with the pre-trained Stable Diffusion (SD) model without fine-tuning on COCO, \ie, the zero-shot setting. 
As shown in the \emph{left three} columns, we experiment with adding ``region description'' and ``region position'' information to the input query in addition to ``image description.'' 
Since T2I models can not understand coordinates, we carefully design positional text descriptions, indicated by ``text'' in the ``region position'' column.
Specifically, we describe a region with one of the three size words (\emph{small, medium, large}), three possible region aspect ratios (\emph{long, square, tall}), and nine possible locations (\emph{top left, top, $\ldots$, bottom right}). The \emph{bottom part} compares the main \modelname~model with other variants fine-tuned with the corresponding input queries.
The \emph{middle three} rows report the results on region control accuracy. For AP and AP$_\text{50}$, we use a DETR ResNet-50 object detector trained on COCO~\cite{carion2020end,DETRRes50} to get the detection results on images generated based on the input texts and boxes from the COCO 2017 val5k set~\cite{lin2014microsoft}.
The ``object accuracy'' column reports the region classification accuracy~\cite{zhao2019image}.
The trained ResNet-101 region classifier~\cite{he2016deep} yields a $71.41\%$ oracle 80-class accuracy on real images. %the 30k val subset.
The \emph{right two} columns report the image generation quality metrics, \ie, SceneFID and FID, which evaluate the region and image visual qualities.

One advantage of \modelname~is its strong region control capability. As shown in the bottom row, \modelname~achieves an AP of $32.0$, which is close to the real image oracle of $36.8$. Despite the careful engineering of positional text words, \modelname$_\text{Position Word}$ only achieves an AP of $2.3$. Similarly, for object region classification, $62.42\%$ of the cropped regions on \modelname-generated images can be correctly classified, compared with $42.02\%$ of \modelname$_\text{Position Word}$.
\modelname~also improves the generated image quality, both at the region and image level. At the region level, \modelname~achieves a SceneFID of $6.51$, indicating strong capabilities in both generating high-fidelity objects and precisely placing them in the queried position.
At the image level, \modelname~improves the FID from $10.44$ to $7.36$ with the region-controlled text input that provides a localized and more detailed image description. We present additional FID comparisons to state-of-the-art conditional image generation methods in Table~\ref{table:sota} (c). 

We show representative qualitative results in Figure~\ref{fig:visucoco}. 
\textbf{(a)} \modelname~can more reliably generate images that involve counting or complex object relationships, \eg, ``five birds'' and ``sitting on a bench.'' 
\textbf{(b)} \modelname~can more easily generate images with unique camera views by controlling the relative position and size of object boxes, \eg, ``a top-down view of a cat'' that T2I models struggle with.
\textbf{(c)} Separating detailed regional descriptions with position tokens also helps \modelname~better understand long queries and reduce attribute leakage, \eg, the color of the clock and person's shirt.

%%%%%%%%%%%%%%%%%%%%%%%%%%%%%%%%%%%%%%%%%%%%%%%%%%%%%%%%%%%%%%%%%%
%%%%%%%%%%%%%%%%%%%%%%%%%%%%%%%%%%%%%%%%
\begin{table}[t]
\centering
\tablestyle{3.5pt}{1.1}
\footnotesize
\begin{tabular}{ l | c c c | c c c }
    \hline
    \multirow{2}{*}{Method} & \multicolumn{3}{c|}{Skill Correctness ($\uparrow$)} & \multicolumn{3}{c}{Object Accuracy ($\uparrow$)} \\
     & Object & Count & Spatial & Object & Count & Spatial  \\
    \hline
    \footnotesize{SD V1.4 Zero-shot} & 97.11 & 59.28 & 48.20 & 35.97 & 22.32 & 13.06 \\
    \modelname$_\text{Image Descr.}$ & 98.23 & 60.40 & 49.11 & 38.92 & 25.79 & 15.17\\
    \modelname$_\text{Position Word}$ & 93.33 & 68.10 & 64.87 & 50.72 & 25.35 & 22.82\\
    \modelname & \textbf{98.51} & \textbf{87.38} & \textbf{82.08} & \textbf{82.30} & \textbf{63.40} & \textbf{67.30} \\
    \hline
\end{tabular}
\caption{Evaluations on the images generated with PaintSkill~\cite{cho2022dall} prompts. We evaluate skill correctness with human judges, and object classification accuracy with the COCO-trained classifier.}
\label{table:paint}
\vspace{-1mm}
\end{table}
%%%%%%%%%%%%%%%%%%%%%%%%%%%%%%%%%%%%%%%%
\noindent\textbf{PaintSkill.}
Table~\ref{table:paint} shows the skill correctness and region control accuracy evaluations on PaintSkill~\cite{cho2022dall}. Skill correctness~\cite{cho2022dall} evaluates if the generated images contain the query-described object type/count/relationship, \ie, the ``object,'' ``count,'' and ``spatial'' subsets. We use human judges to obtain the skill correctness accuracy.
For region control, we use object classification accuracy to evaluate if the model follows those arbitrarily shaped and located object queries. We reuse the COCO region classifier introduced in Table~\ref{table:coco}.

Based on the human evaluation for ``skill correctness,'' $87.38\%$ and $82.08\%$ of \modelname-generated images have the correct object count and spatial relationship (``count'' and ``spatial''), which is $+19.28\%$ and $+17.21\%$ more accurate than \modelname$_\text{Position Word}$, and $+26.98\%$ and $+32.97\%$ higher than the T2I model with image description only. 
The skill correctness improvements suggest that region-control text inputs could be an effective interface to help T2I models more reliably generate user-specified scenes. The object accuracy evaluation makes the criteria more strict by requiring the model to follow the exact input region positions, in addition to skills. ``\modelname'' achieves a strong region control accuracy of $63.40\%$ and $67.30\%$ on count and skill subsets, surpassing ``\modelname$_\text{Position Word}$'' by $+38.05\%$ and $+44.48\%$.

PaintSkill contains input queries with randomly assigned object types, locations, and shapes. Because of the minimal constraints, many queries describe challenging scenes that appear less frequently in real life. We observe that \modelname~not only precisely follows position queries, but also fits objects and their surroundings naturally, indicating an understanding of object properties.
In Figure~\ref{fig:visupaint} (a), the three buses with different aspect ratios each have their unique viewing angle and direction, such that the object ``bus'' fits tightly with the given region. More interestingly, the directions of each bus go nicely with the road, making the image look real to humans. Figure~\ref{fig:visupaint} (b) shows challenging cases that require drawing two less commonly co-occurred objects into the same image. \modelname~correctly fits ``bed'' and ``fire hydrant,'' ``boat'' and ``bus'' into the given region. 
More impressively, \modelname~can create a scene that makes the generated image look plausible, \eg, ``looking through a window with a bed indoors,'' with the commonsense knowledge that ``bed'' is usually indoor while ``fire hydrant'' is usually outdoor.
The randomly assigned region categories can also lead to objects with unusual relative sizes, \eg, the bag that is larger than the airplane in Figure~\ref{fig:visupaint} (c). \modelname~shows an understanding of image perspectives by placing smaller objects such as ``backpack'' and ``dog'' near the camera position.

%%%%%%%%%%%%%%%%%%%%%%%%%%%%%%%%%%%%%%%%
\begin{figure}[t]
\centering
\includegraphics[width=.47\textwidth]{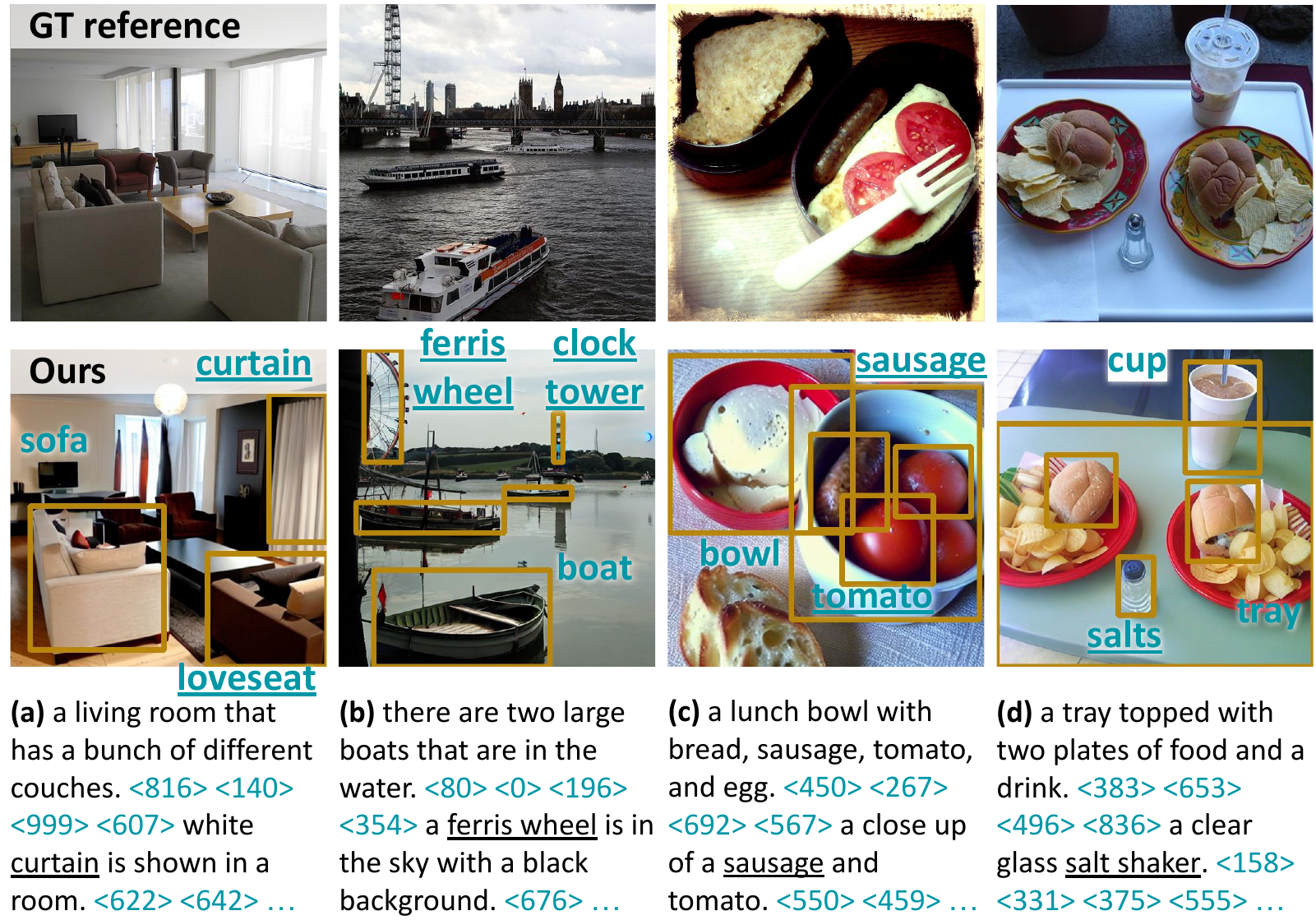}
\vspace{0.05in}
    \caption{
    Qualitative results on LVIS~\cite{gupta2019lvis}. \modelname~can understand open-vocabulary regional descriptions, including keywords such as ``curtain,'' ``ferris wheel,'' ``sausage,'' and ``salt shaker.''
	}
\label{fig:visulvis}
\end{figure}
%%%%%%%%%%%%%%%%%%%%%%%%%%%%%%%%%%%%%%%%%%%%%%%%%%%%%%%%%%%%%%%%%%

%%%%%%%%%%%%%%%%%%%%%%%%%%%%%%%%%%%%%%%%
\begin{figure*}[t]
% \vspace{-0.04in}
\centering
\includegraphics[width=.95\textwidth]{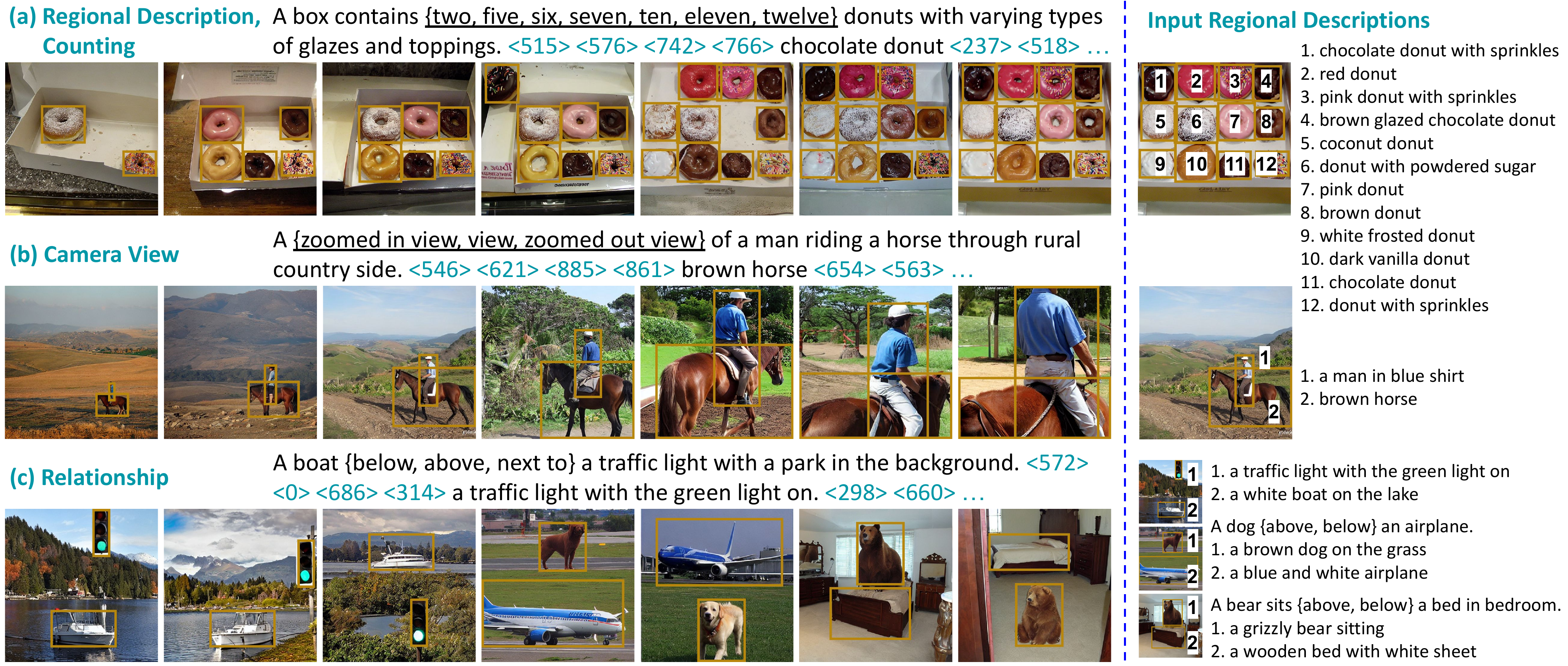}
    \caption{
    Qualitative results of \modelname-generated images with manually designed input queries.
	}
\vspace{3pt}
\label{fig:visu}
\end{figure*}
%%%%%%%%%%%%%%%%%%%%%%%%%%%%%%%%%%%%%%%%%%%%%%%%%%%%%%%%%%%%%%%%%%
\noindent\textbf{LVIS.}
Table~\ref{table:lvis} reports the T2I generation results with out-of-vocabulary regional entities. We observe that \modelname~can understand open-vocabulary regional descriptions, by transferring the open-vocab capability learned from large-scale T2I pre-training to regional descriptions. 
\modelname~achieves the best SceneFID and object classification accuracy over the 1,203 LVIS classes of $10.08$ and $23.42\%$. The results show that the \modelname~position tokens can be used with open-vocabulary regional descriptions, despite being trained on COCO with 80 object types. Figure~\ref{fig:visulvis} shows examples of generating objects that are not annotated in COCO, \eg, ``curtain'' and ``loveseat'' in (a), ``ferris wheel'' and ``clock tower'' in (b), ``sausage'' and ``tomato'' in (c), ``salts'' in (d).

%%%%%%%%%%%%%%%%%%%%%%%%%%%%%%%%%%%%%%%%
\begin{table}[t]\small
\centering
\tablestyle{7pt}{1.1}
\footnotesize
\begin{tabular}{ l | c c c }
    \hline
    Method & \footnotesize{Object Acc. ($\uparrow$)} & \footnotesize{SceneFID ($\downarrow$)} & \footnotesize{FID ($\downarrow$)} \\
    \hline
    Real Images & \textcolor{gray}{42.00} & \textcolor{gray}{-} & \textcolor{gray}{-} \\
    \hline
    \footnotesize{SD V1.4 Zero-shot} & 7.88 & 40.62 & 23.74 \\
    \modelname$_\text{Image Descr.}$ & 9.82 & 28.95 & 20.87 \\
    \modelname$_\text{Region Descr.}$ & 11.08 & 28.15 & 17.96 \\
    \modelname$_\text{Position Word}$ & 16.60 & 20.27 & 17.80 \\
    \modelname & \textbf{23.42} & \textbf{10.08} & \textbf{17.73} \\
    \hline
\end{tabular}
\caption{Evaluations on the images generated with the 4,809 LVIS validation samples~\cite{gupta2019lvis} from COCO val2017. The object classification is conducted over the 1,203 LVIS classes.}
\label{table:lvis}
\vspace{6pt}
\end{table}
%%%%%%%%%%%%%%%%%%%%%%%%%%%%%%%%%%%%%%%%

%%%%%%%%%%%%%%%%%%%%%%%%%%%%%%%%%%%%%%%%
%%%%%%%%%%%%%%%%%%%%%%%%%%%%%%%%%%%%%%%%
\begin{table}[t]
\centering
\tablestyle{4pt}{1.1}
\footnotesize
\begin{tabular}{ l | c c c | c c c }
    \hline
    \multirow{2}{*}{Method} &
    \multicolumn{3}{c|}{COCO} & \multicolumn{3}{c}{LVIS}\\
     & \footnotesize{Acc.} & \footnotesize{SceneFID} & \footnotesize{FID} & \footnotesize{Acc.} & \footnotesize{SceneFID} & \footnotesize{FID} \\
    \hline
    Real Images & \textcolor{gray}{74.41} & - & - & \textcolor{gray}{42.00} & - & - \\
    \hline
    \modelname$_\text{OD Label}$ & \textbf{69.70} & 8.07 & 9.08 & 22.79 & 13.98 & 23.06 \\
    \modelname & 62.42 & \textbf{6.51} & \textbf{7.36} & \textbf{23.42} & \textbf{10.08} & \textbf{17.73} \\
    \hline
\end{tabular}
\caption{Analyses on using open-ended texts (\modelname) \vs constrained object labels (\modelname$_\text{OD Label}$) as the regional description.}
\label{table:descrtype}
\end{table}
%%%%%%%%%%%%%%%%%%%%%%%%%%%%%%%%%%%%%%%%
%%%%%%%%%%%%%%%%%%%%%%%%%%%%%%%%%%%%%%%%
\begin{table*}[t]\small
\resizebox{.95\textwidth}{!}
{
\begin{minipage}{.35\textwidth}
  \centering
  \includegraphics[height=.8\textwidth]{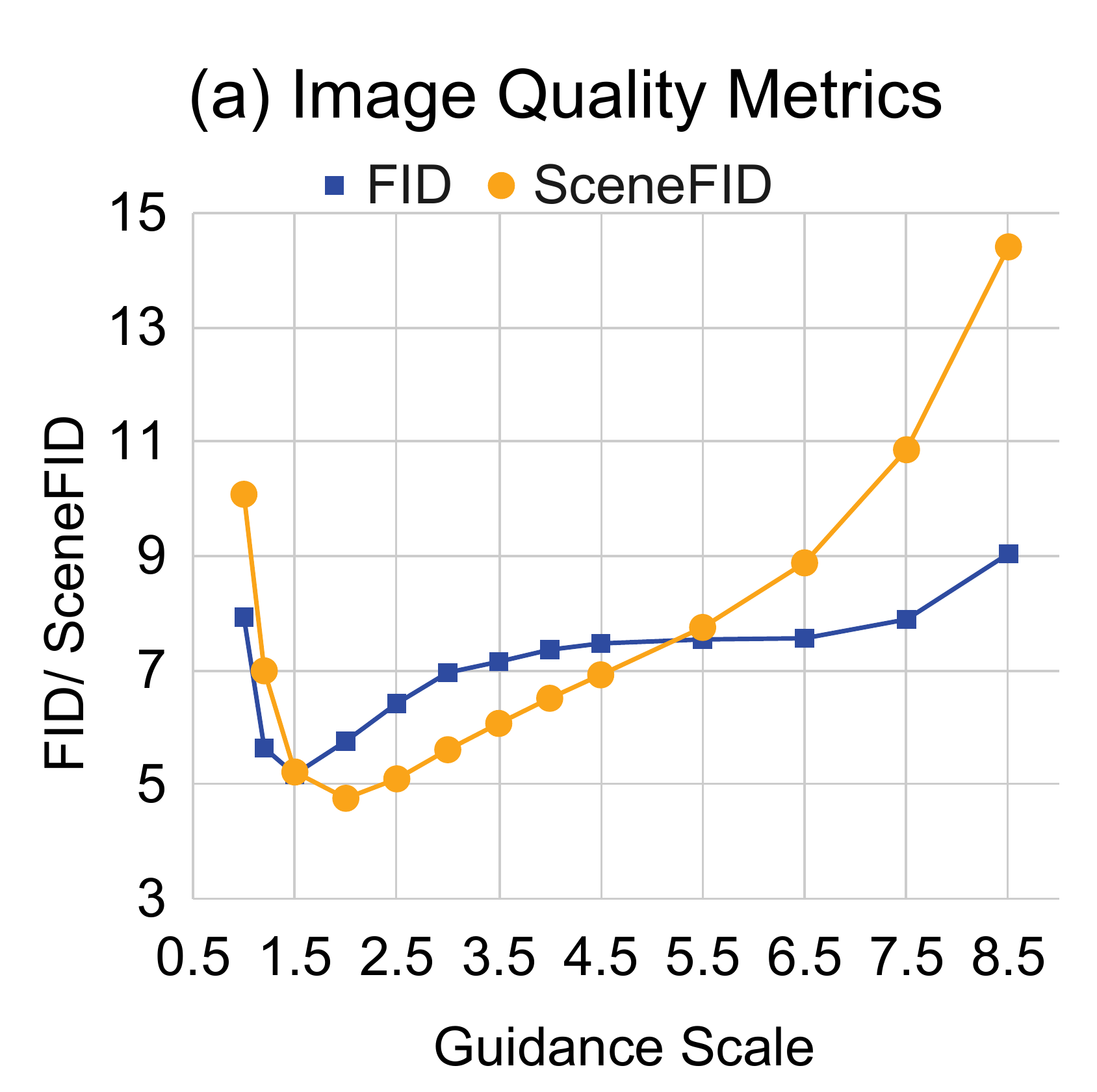}
\end{minipage}
\quad
\begin{minipage}{.35\textwidth}
  \centering
  \includegraphics[height=.8\textwidth]{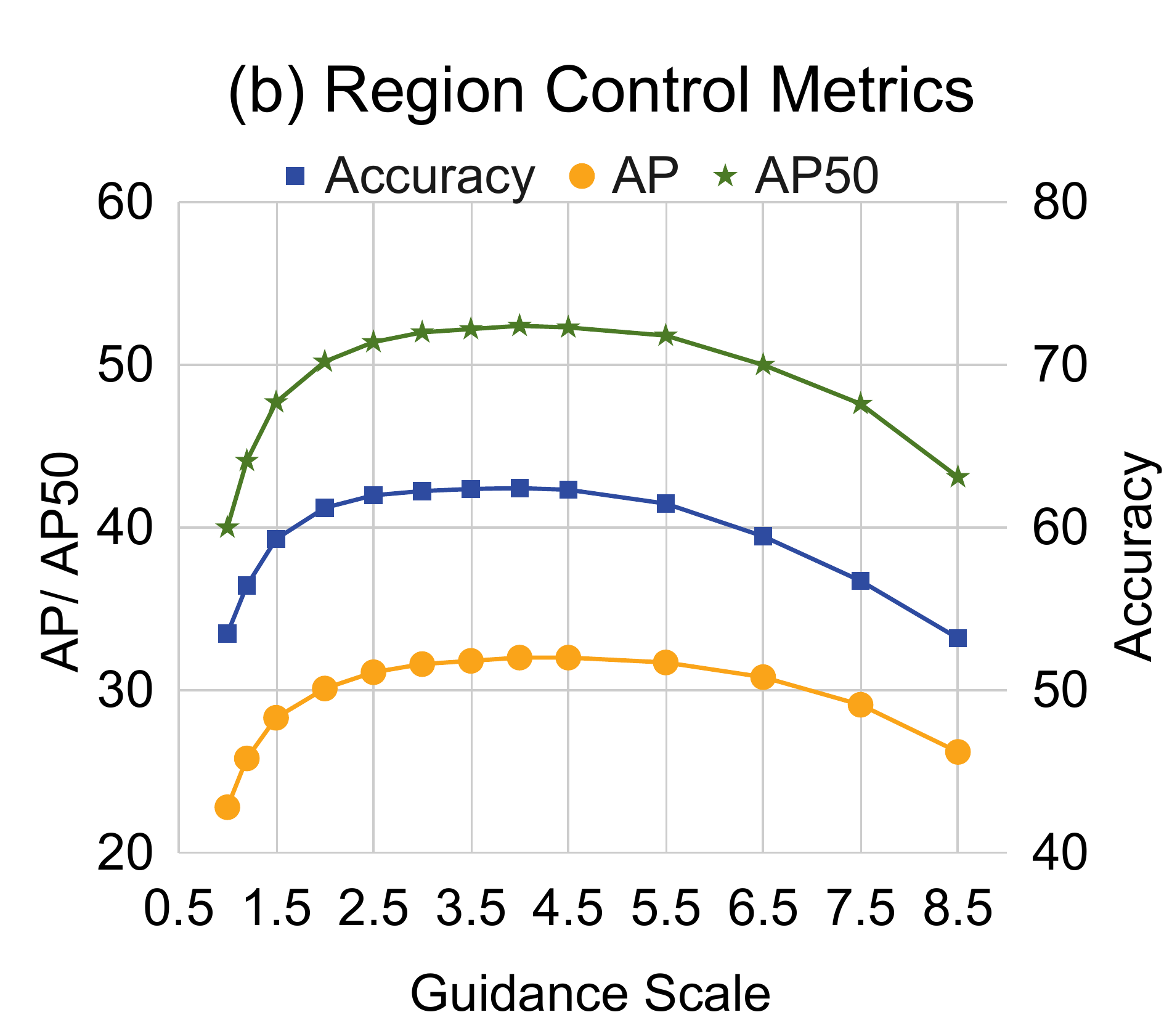}
\end{minipage}
\quad
\begin{minipage}{.4\textwidth}
    \centering
    \tablestyle{10pt}{1.0}
    \begin{tabular}{ l | c }
        \hline
        (c) Method & FID ($\downarrow$) \\
        \hline
        Random Train Images~\cite{gafni2022make} & 2.47 \\
        Retrieval Baseline~\cite{yu2022scaling} & 6.82 \\
        \hline
        XMC-GAN~\cite{zhang2021cross} & 9.33 \\
        CogView2~\cite{ding2022cogview2} & 17.7 \\
        LAFITE~\cite{zhou2022towards} & 8.12 \\
        Make-A-Scene~\cite{gafni2022make} & 7.55 \\
        Parti~\cite{yu2022scaling} & 3.22 \\
        \hline
        \modelname$_\text{Image Descr.}$ & 6.98 \\
        \modelname$_\text{Position Word}$ & 5.98\\
        \modelname & 5.18 \\
        \hline
    \end{tabular}
    \vspace{-0.0in}
\end{minipage}%
}
\caption{\textbf{(a,b)} Analyses of different guidance scales' influences~\cite{ho2022classifier} on image quality and region control accuracy. \textbf{(c)} Comparison with previous T2I works on the COCO (2014) validation 30k subset~\cite{lin2014microsoft,ramesh2021zero,yu2022scaling,wang2022ofa}) in the fine-tuned setting.}
\label{table:sota}
% \vspace{-0.02in}
\end{table*}
%%%%%%%%%%%%%%%%%%%%%%%%%%%%%%%%%%%%%%%%

%%%%%%%%%%%%%%%%%%%%%%%%%%%%%%%%%%%%%%%%
\noindent\textbf{Qualitative results.}
We next qualitatively show \modelname's other capabilities with manually designed input queries. Figure~\ref{fig:teaser} (a) shows examples of arbitrary object manipulation and regional description control.
As shown in the ``bus'' example, \modelname~will automatically adjust the object viewing (from side to front) and type (from single- to double-deck) to reasonably fit the region constraint, indicating the knowledge about object ``bus.'' \modelname~can also understand the free-form regional text and generate ``cats'' in the specified region with different attributes, \eg, ``wearing a red hat,'' ``pink,'' ``sleeping,'' \etc.
Figure~\ref{fig:visu} {(a)} shows an example of generating images with different object counts. \modelname's region control provides a strong tool for generating the exact object count, optionally with extra regional texts describing each object.
Figure~\ref{fig:visu} {(b)} shows how we can use the box size to control the camera view, \eg, the precise control of the exact zoom-in ratio.
Figure~\ref{fig:visu} {(c)} presents additional examples of images with unusual object relationships.

% %%%%%%%%%%%%%%%%%%%%%%%%%%%%%%%%%%%%%%%%
\subsection{Analysis}
\label{sec:ana}
\noindent\textbf{Regional descriptions.}
Alternative to the open-ended free-form texts, regional descriptions can be object indexes from a constrained category set, as the setup in layout-to-image generation~\cite{zhao2019image,sun2019image,li2020bachgan,li2021image,frolov2021attrlostgan}.
Table~\ref{table:descrtype} compares \modelname~with \modelname$_\text{OD Label}$ on COCO~\cite{lin2014microsoft} and LVIS~\cite{gupta2019lvis}.
The leftmost ``accuracy'' column on COCO shows the major advantage of \modelname$_\text{OD Label}$, \ie, when fine-tuned and tested with the same regional object vocabulary, \modelname$_\text{OD Label}$ is $+7.28\%$ higher in region control accuracy, compared with \modelname. However, the closed-vocabulary OD labels bring two \emph{disadvantages}. \emph{First}, the position tokens in \modelname$_\text{OD Label}$ tend to only work with the seen vocabulary, \ie, the 80 COCO categories. When evaluated on other datasets such as LVIS or open-world use cases, the region control performance drops significantly, as shown in the ``accuracy'' column on LVIS. \emph{Second}, \modelname$_\text{OD Label}$ only works well with constrained object labels, which fail to provide detailed regional descriptions, such as attributes and object relationships. Therefore, \modelname$_\text{OD Label}$ helps less in generating high-fidelity images, with FID $1.72$ and $5.33$ worse than \modelname~on COCO and LVIS.
Given the aforementioned limitations, we use the open-ended free-form regional descriptions in \modelname.

\noindent\textbf{Guidance scale and T2I SOTA comparison.}
Table~\ref{table:sota} (a,b) examines how different classifier-free guidance scales~\cite{ho2022classifier} influence region control accuracy and image generation quality on the COCO 2014 validation subset~\cite{lin2014microsoft,ramesh2021zero,yu2022scaling,wang2022ofa}. We empirically observe that scale of $1.5$ yields the best image quality, and a slightly larger scale of $4.0$ provides the best region control performance.
Table~\ref{table:sota} (c) compares \modelname~with the state-of-the-art T2I methods in the fine-tuned setting. 
We reduce the guidance scale from the $4.0$ in Table~\ref{table:coco} to $1.5$ for a fair comparison. We do not use any image-text contrastive models for results re-ranking. 
\modelname~achieves an FID of $5.18$, compared with $6.98$ when we fine-tune Stable Diffusion with COCO T2I data without regional description. 
\modelname~also outperforms the real image retrieval baseline~\cite{yu2022scaling} and most prior studies~\cite{zhang2021cross,ding2022cogview2,zhou2022towards,gafni2022make}.

\noindent\textbf{Limitations.}
Our method has several limitations. 
First, \modelname~might generate lower-quality images when the input query becomes too challenging, \eg, the unusual giant ``dog'' in Figure~\ref{fig:visu} (c).
Second, for evaluation purposes, we train \modelname~on the COCO training set. Despite preserving the open-vocabulary capability shown on LVIS, the generated image style does bias towards COCO. This limitation can potentially be alleviated by conducting the same \modelname~fine-tuning on a small subset of pre-training data~\cite{schuhmann2022laion} used by the same T2I model~\cite{rombach2022high}. We show this \modelname~variant in the supplementary material. 
% Third, \modelname~takes user-generated position tokens. Instead, position tokens can also be predicted from text~\cite{hong2018inferring,li2019object}, potentially enabling more applications.
Finally, \modelname~builds upon large-scale pre-trained T2I models such as Stable Diffusion~\cite{rombach2022high} and shares similar possible generation biases.

\vspace{-3mm}
\section{Conclusion}
\vspace{-1mm}
We have presented \modelname~that extends a pre-trained T2I model for region-controlled T2I generation. Our introduced position token allows the precise specification of open-ended regional descriptions on arbitrary image regions, leading to an effective new interface of region-controlled text input. We show that \modelname~can help T2I generation in challenging cases, \eg, when the input query is complicated with detailed regional attributes or describes an unusual scene.
Experiments validate \modelname's effectiveness on both region control accuracy and image generation quality.

\clearpage
% %%%%%%%%% TITLE - PLEASE UPDATE
\title{ReCo: Region-Controlled Text-to-Image Generation\\(Supplementary Material)}

\author{}

% % % \maketitle
\twocolumn[{\renewcommand\twocolumn[1][]{#1}\maketitle
\centering
\small
\vspace{-2em}
\includegraphics[width=0.8\textwidth]{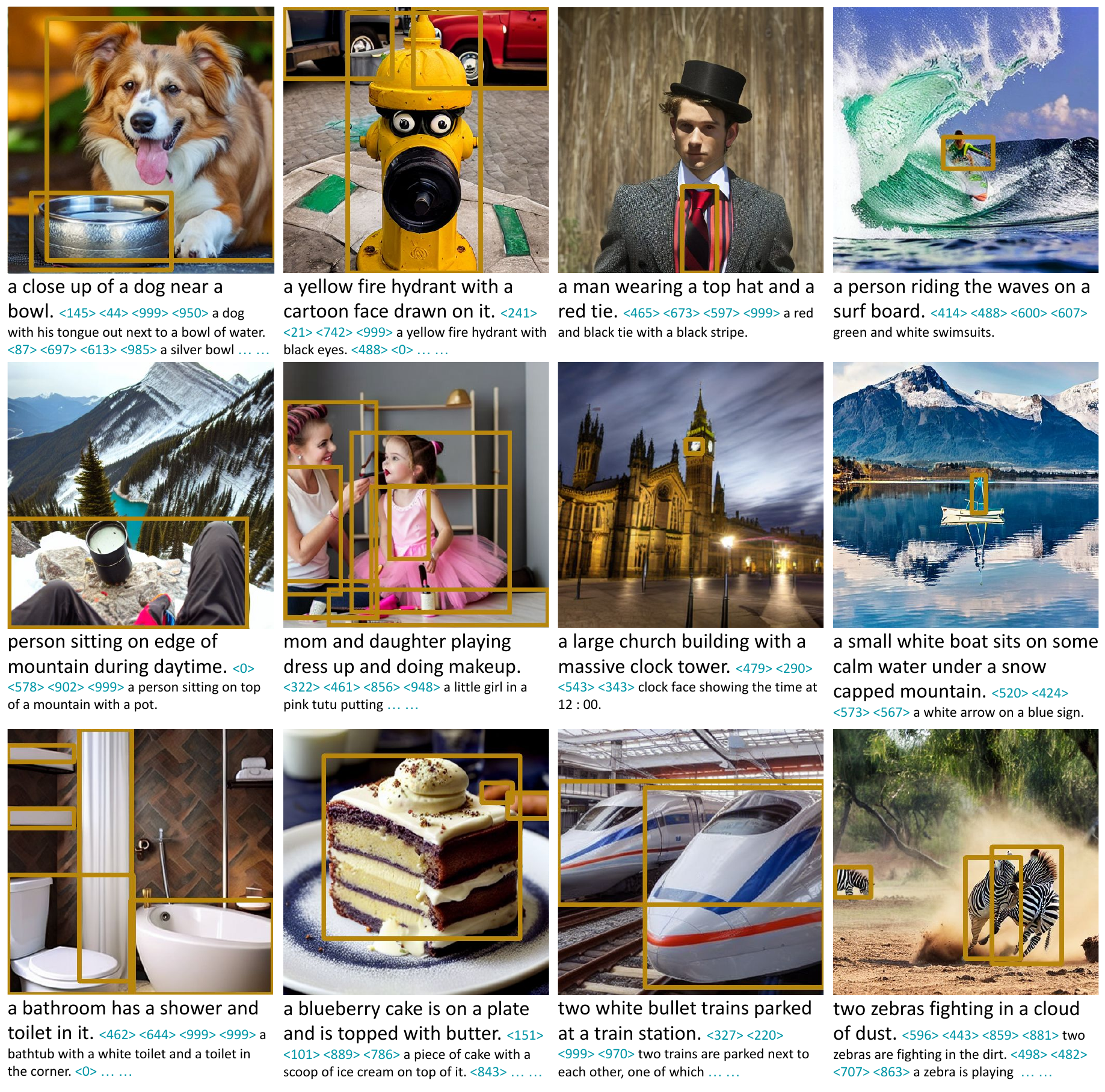} \\
\vspace{1.0em}
\captionof{figure}{
Example images generated by  \modelname$_\text{LAION}$. More examples are in Figures~\ref{fig:laionlvis_zoom}, \ref{fig:laionlaion_zoom}.
}
\label{fig:laion_teaser}
\vspace{4.0em}}]

\appendix

%%%%%%%%%%%%%%%%%%%%%%%%%%%%%%%%%%%%%%%%
\begin{figure*}[t]
\centering
\includegraphics[width=.95\textwidth]{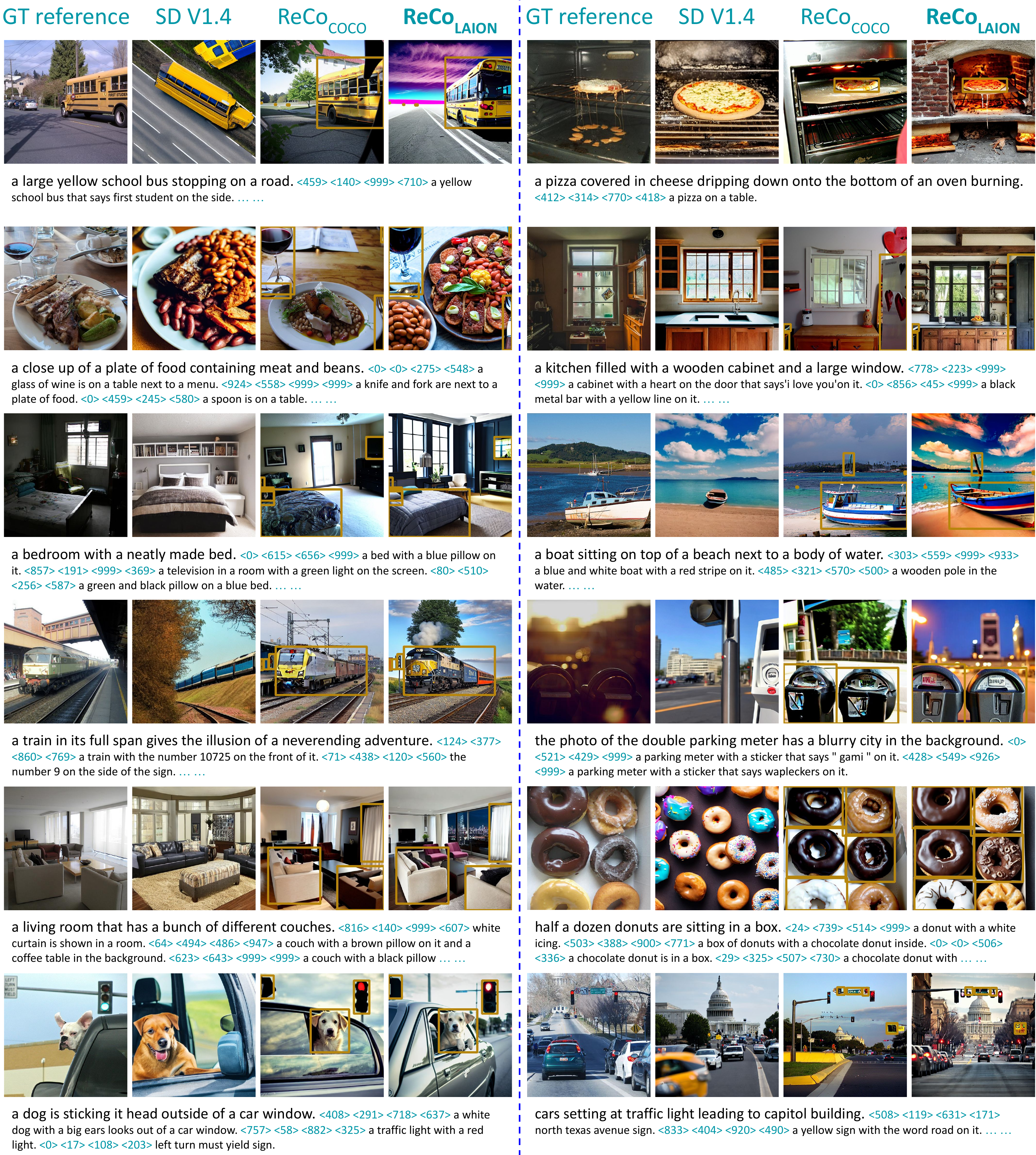}
    \caption{
    Qualitative results on LVIS~\cite{gupta2019lvis}. Zoomed-in version is in Figure~\ref{fig:laionlvis_zoom}.
	}
\label{fig:laionlvis}
\end{figure*}
%%%%%%%%%%%%%%%%%%%%%%%%%%%%%%%%%%%%%%%%%%%%%%%%%%%%%%%%%%%%%%%%%%
%%%%%%%%%%%%%%%%%%%%%%%%%%%%%%%%%%%%%%%%
\begin{figure*}[t]
\centering
\includegraphics[width=.95\textwidth]{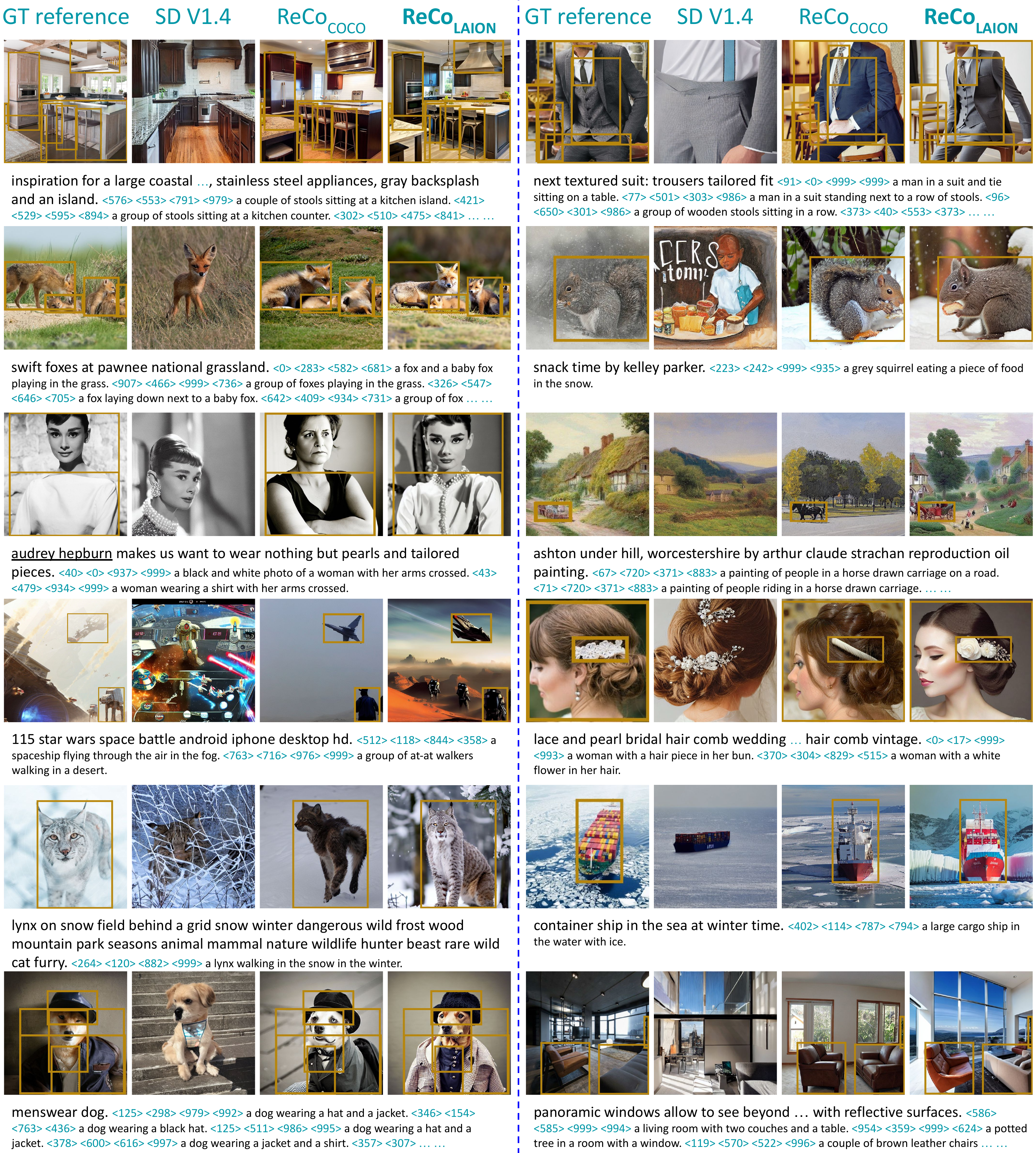}
    \caption{
    Qualitative results on prompts selected from LAION-Aesthetics~\cite{schuhmann2022laion}. Zoomed-in version is in Figure~\ref{fig:laionlaion_zoom}.
	}
\label{fig:laionlaion}
\end{figure*}
%%%%%%%%%%%%%%%%%%%%%%%%%%%%%%%%%%%%%%%%%%%%%%%%%%%%%%%%%%%%%%%%%%
%%%%%%%%%%%%%%%%%%%%%%%%%%%%%%%%%%%%%%%%
\begin{figure*}[t]
\centering
\includegraphics[width=.95\textwidth]{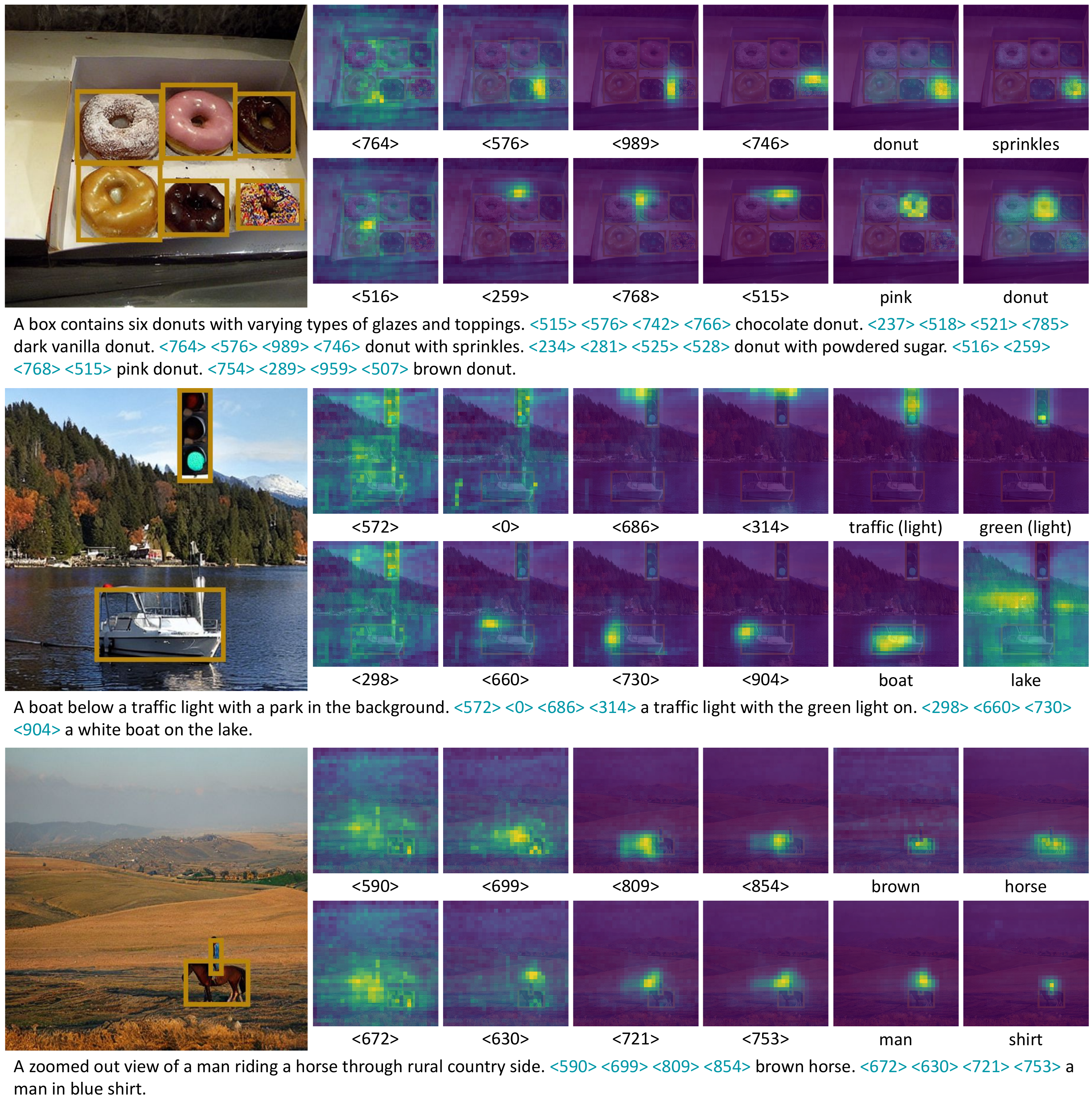}
    \caption{
    Averaged \modelname$_\text{COCO}$~cross-attention maps between visual latent and text embedding (on both text and position tokens).
	}
\vspace{3pt}
\label{fig:attn}
\vspace{5pt}
\end{figure*}
%%%%%%%%%%%%%%%%%%%%%%%%%%%%%%%%%%%%%%%%%%%%%%%%%%%%%%%%%%%%%%%%%%

%%%%%%%%%%%%%%%%%%%%%%%%%%%%%%%%%%%%%%%%
\section{\modelname~with LAION data}
In the main paper, we focus on the \modelname~model trained on COCO (\modelname$_\text{COCO}$) to standardize the evaluation process. In this section, we present \modelname$_\text{LAION}$ that conducts the same \modelname~fine-tuning on a small subset of the LAION dataset~\cite{schuhmann2022laion} used by the pre-trained SD model~\cite{rombach2022high}. Figure~\ref{fig:laion_teaser} shows selected \modelname$_\text{LAION}$-generated image samples.

%%%%%%%%%%%%%%%%%%%%%%%%%%%%%%%%%%%%%%%%
\noindent\textbf{Training setup.}
Instead of using the 414K image-text pairs (83K images) from the COCO 2014 training set, we randomly sample 100K images from the LAION-Aesthetics dataset\footnote{We use the first 100K samples with an aesthetics score of 6 or higher following the index in \url{https://huggingface.co/datasets/ChristophSchuhmann/improved_aesthetics_6plus}.}. We take the Detic object detector~\cite{zhou2022detecting} to generate the object region predictions. We use a confidence threshold of $0.5$ and filter out small boxes with a size smaller than $0.03\times W\times H$. Following the setting for \modelname$_\text{COCO}$, we feed all cropped regions to the pre-trained GIT captioning model~\cite{wang2022git} for regional descriptions. We fine-tune \modelname~for 10,000 steps with the same training and inference settings introduced in the main paper.

%%%%%%%%%%%%%%%%%%%%%%%%%%%%%%%%%%%%%%%%%%%%%%%%%%%%%%%%%%%%%%%%%%
\noindent\textbf{Qualitative results.}
Figure~\ref{fig:laionlvis} shows qualitative results on LVIS~\cite{gupta2019lvis}. Both \modelname$_\text{COCO}$ and \modelname$_\text{LAION}$ show strong region-controlled T2I generation capabilities. Compared with \modelname$_\text{COCO}$, \modelname$_\text{LAION}$-generated images have better image aesthetic scores, thanks to the high-aesthetic fine-tuning data from LAION~\cite{schuhmann2022laion}.

Figure~\ref{fig:laionlaion} shows qualitative results on LAION-Aesthetics. We run T2I inference on 3K samples indexed after the first 100K samples used for \modelname~fine-tuning. \modelname$_\text{LAION}$ can preserve the pre-trained SD's capabilities of understanding celebrities, art styles, and open-vocabulary descriptions, and meanwhile extend SD with the appealing new ability of region-controlled T2I generation. 

%%%%%%%%%%%%%%%%%%%%%%%%%%%%%%%%%%%%%%%%%%%%%%%%%%%%%%%%%%%%%%%%%%
%%%%%%%%%%%%%%%%%%%%%%%%%%%%%%%%%%%%%%%%
\begin{table}[t]\small
\centering
\tablestyle{7pt}{1.1}
\footnotesize
\begin{tabular}{ l c | c c c }
    \hline
    \multirow{2}{*}{Method} & COCO & Object & \multirow{2}{*}{SceneFID ($\downarrow$)} & \multirow{2}{*}{FID ($\downarrow$)} \\
     & Image & Acc. ($\uparrow$) &  &  \\
    \hline
    Real Images & - & \textcolor{gray}{42.00} & \textcolor{gray}{-} & \textcolor{gray}{-} \\
    \hline
    \footnotesize{SD V1.4} & \xmark & 7.88 & 40.62 & 23.74 \\
    \modelname$_\text{COCO}$ & \cmark & \textbf{23.42} & \textbf{10.08} & \textbf{17.73} \\
    \modelname$_\text{LAION}$ & \xmark & {19.38} & {19.48} & {21.99} \\
    \hline
\end{tabular}
\caption{Evaluations on the images generated with the 4,809 LVIS validation samples~\cite{gupta2019lvis} from COCO val2017. The object classification is conducted over the 1,203 LVIS classes.}
\label{table:laion_lvis}
\vspace{4pt}
\end{table}
%%%%%%%%%%%%%%%%%%%%%%%%%%%%%%%%%%%%%%%%

\noindent\textbf{Quantitative results.}
Table~\ref{table:laion_lvis} compares \modelname$_\text{LAION}$ with \modelname$_\text{COCO}$ on LVIS~\cite{gupta2019lvis}. The ``COCO Image'' column indicates if the COCO image style is seen during \modelname~fine-tuning. Automatic metrics show that \modelname$_\text{COCO}$ achieves better region control accuracy and image FID. For region control, COCO ground-truth boxes provide a cleaner region specification than Detic-predicted boxes, thus benefiting the controlling accuracy. For the FID evaluation, \modelname$_\text{COCO}$ has seen COCO images during \modelname~training, leading to better FID scores. Qualitatively, \modelname$_\text{LAION}$-generated images show comparable, if not better visual qualities than \modelname$_\text{COCO}$. Overall, both \modelname~model variants significantly outperform the original SD model in both region control accuracy and image generation quality.

%%%%%%%%%%%%%%%%%%%%%%%%%%%%%%%%%%%%%%%%
\section{Position Token Cross-Attention}
%%%%%%%%%%%%%%%%%%%%%%%%%%%%%%%%%%%%%%%%
To help interpret how the introduced position tokens operate, Figure~\ref{fig:attn} visualizes the cross-attention maps between the visual latent $z$ and token embedding $\tau_{\theta}(y(P,T))$. We show the averaged attention maps across all diffusion steps and U-Net blocks. We empirically observe that the four position tokens for each region help the model to gradually localize the specified area by attending to the corner or edge positions of the box region. The position tokens help text tokens to localize and focus on the detailed regional descriptions, \eg, the ``green light'' in the ``traffic light.''

%%%%%%%%%%%%%%%%%%%%%%%%%%%%%%%%%%%%%%%%
\begin{figure*}[t]
\centering
\includegraphics[width=.95\textwidth]{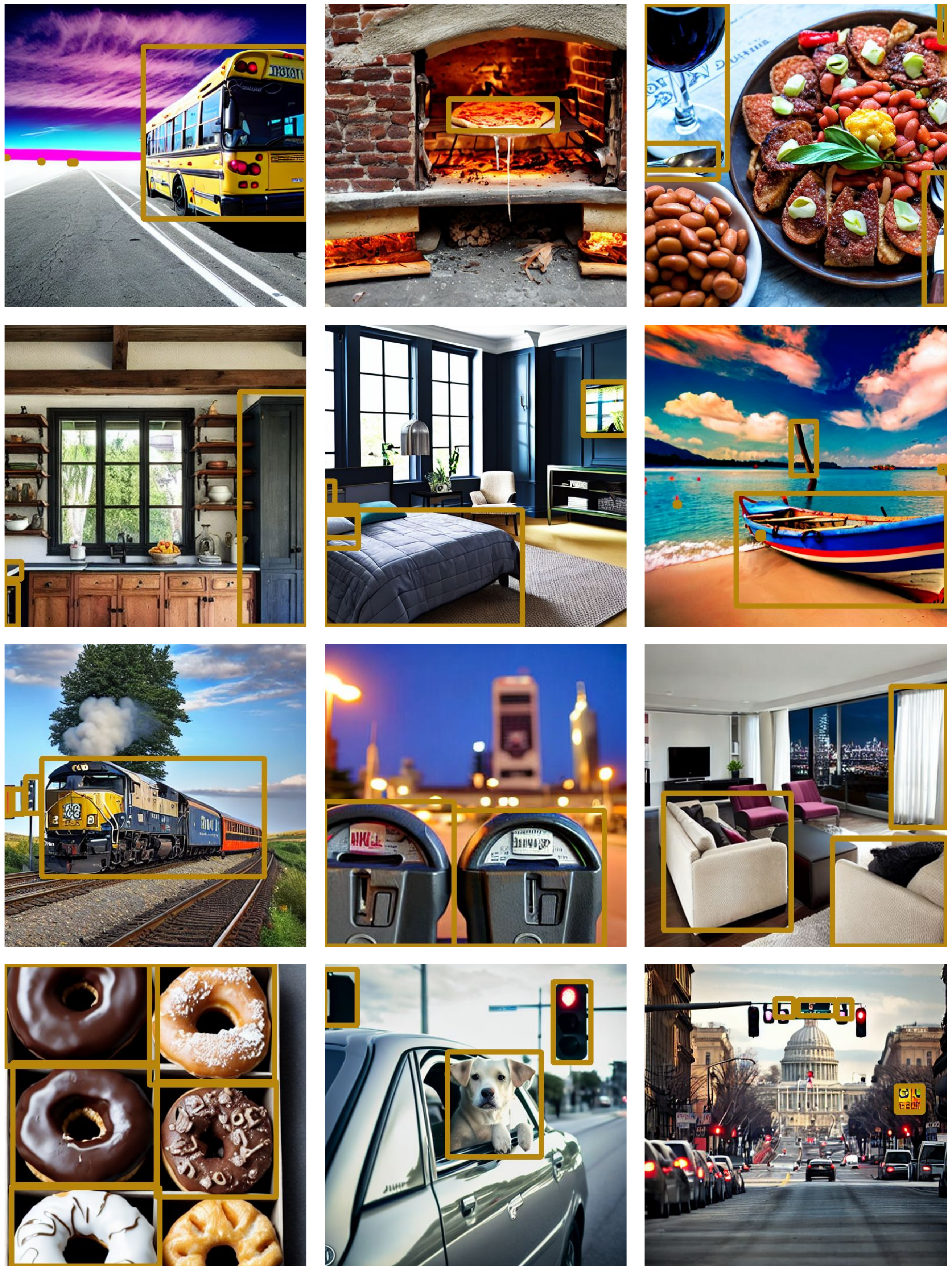}
    \caption{
    Zooming in \modelname$_\text{LAION}$-generated images shown in Figure~\ref{fig:laionlvis}.
	}
\label{fig:laionlvis_zoom}
\end{figure*}
%%%%%%%%%%%%%%%%%%%%%%%%%%%%%%%%%%%%%%%%%%%%%%%%%%%%%%%%%%%%%%%%%%
%%%%%%%%%%%%%%%%%%%%%%%%%%%%%%%%%%%%%%%%
\begin{figure*}[t]
\centering
\includegraphics[width=.95\textwidth]{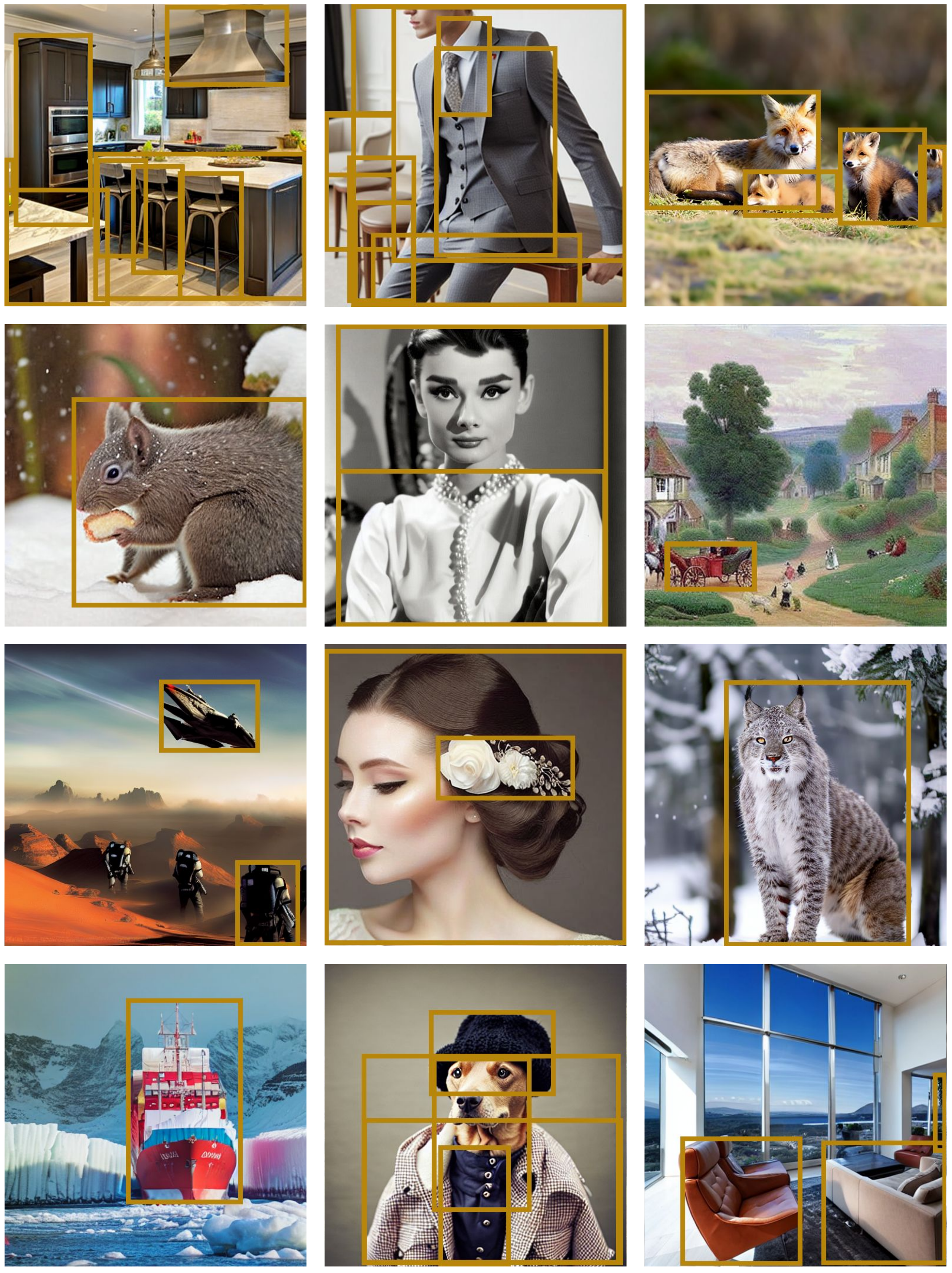}
    \caption{
    Zooming in \modelname$_\text{LAION}$-generated images shown in Figure~\ref{fig:laionlaion}.
	}
\label{fig:laionlaion_zoom}
\end{figure*}
%%%%%%%%%%%%%%%%%%%%%%%%%%%%%%%%%%%%%%%%%%%%%%%%%%%%%%%%%%%%%%%%%%

\clearpage
%%%%%%%%% REFERENCES
% {\small
{\small
% {\footnotesize
\vspace{-1pt}
\subsection*{Acknowledgment}
\vspace{-2pt}
We would like to thank Lin Liang and Faisal Ahmed for their help on human evaluation and data preparation, and Jaemin Cho and Xiaowei Hu for their helpful discussions. We would like to thank Yumao Lu and Xuedong Huang for their support.

\bibliographystyle{ieee_fullname}
\bibliography{egbib}

\begin{thebibliography}{10}\itemsep=-1pt

\bibitem{DETRRes50}
Nicolas Carion, Francisco Massa, Gabriel Synnaeve, Nicolas Usunier, Alexander
  Kirillov, and Sergey Zagoruyko.
\newblock Detr res50 checkpoint from the official detr repo.
\newblock In {\em https://dl.fbaipublicfiles.com/detr/detr-r50-e632da11.pth},
  2020.

\bibitem{carion2020end}
Nicolas Carion, Francisco Massa, Gabriel Synnaeve, Nicolas Usunier, Alexander
  Kirillov, and Sergey Zagoruyko.
\newblock End-to-end object detection with transformers.
\newblock In {\em ECCV}, 2020.

\bibitem{chen2021pix2seq}
Ting Chen, Saurabh Saxena, Lala Li, David~J Fleet, and Geoffrey Hinton.
\newblock Pix2seq: A language modeling framework for object detection.
\newblock In {\em ICLR}, 2022.

\bibitem{chen2015microsoft}
Xinlei Chen, Hao Fang, Tsung-Yi Lin, Ramakrishna Vedantam, Saurabh Gupta, Piotr
  Doll{\'a}r, and C~Lawrence Zitnick.
\newblock Microsoft coco captions: Data collection and evaluation server.
\newblock {\em arXiv preprint arXiv:1504.00325}, 2015.

\bibitem{cho2022dall}
Jaemin Cho, Abhay Zala, and Mohit Bansal.
\newblock Dall-eval: Probing the reasoning skills and social biases of
  text-to-image generative transformers.
\newblock {\em arXiv preprint arXiv:2202.04053}, 2022.

\bibitem{ding2022cogview2}
Ming Ding, Wendi Zheng, Wenyi Hong, and Jie Tang.
\newblock Cogview2: Faster and better text-to-image generation via hierarchical
  transformers.
\newblock {\em arXiv preprint arXiv:2204.14217}, 2022.

\bibitem{fan2022frido}
Wan-Cyuan Fan, Yen-Chun Chen, DongDong Chen, Yu Cheng, Lu Yuan, and
  Yu-Chiang~Frank Wang.
\newblock Frido: Feature pyramid diffusion for complex scene image synthesis.
\newblock {\em arXiv preprint arXiv:2208.13753}, 2022.

\bibitem{frolov2021attrlostgan}
Stanislav Frolov, Avneesh Sharma, J{\"o}rn Hees, Tushar Karayil, Federico Raue,
  and Andreas Dengel.
\newblock Attrlostgan: attribute controlled image synthesis from reconfigurable
  layout and style.
\newblock In {\em DAGM German Conference on Pattern Recognition}, pages
  361--375. Springer, 2021.

\bibitem{gafni2022make}
Oran Gafni, Adam Polyak, Oron Ashual, Shelly Sheynin, Devi Parikh, and Yaniv
  Taigman.
\newblock Make-a-scene: Scene-based text-to-image generation with human priors.
\newblock {\em arXiv preprint arXiv:2203.13131}, 2022.

\bibitem{gupta2019lvis}
Agrim Gupta, Piotr Dollar, and Ross Girshick.
\newblock Lvis: A dataset for large vocabulary instance segmentation.
\newblock In {\em CVPR}, 2019.

\bibitem{he2016deep}
Kaiming He, Xiangyu Zhang, Shaoqing Ren, and Jian Sun.
\newblock Deep residual learning for image recognition.
\newblock In {\em CVPR}, 2016.

\bibitem{heusel2017gans}
Martin Heusel, Hubert Ramsauer, Thomas Unterthiner, Bernhard Nessler, and Sepp
  Hochreiter.
\newblock Gans trained by a two time-scale update rule converge to a local nash
  equilibrium.
\newblock {\em Advances in neural information processing systems}, 30, 2017.

\bibitem{hinz2019generating}
Tobias Hinz, Stefan Heinrich, and Stefan Wermter.
\newblock Generating multiple objects at spatially distinct locations.
\newblock {\em arXiv preprint arXiv:1901.00686}, 2019.

\bibitem{ho2022classifier}
Jonathan Ho and Tim Salimans.
\newblock Classifier-free diffusion guidance.
\newblock {\em arXiv preprint arXiv:2207.12598}, 2022.

\bibitem{hong2018inferring}
Seunghoon Hong, Dingdong Yang, Jongwook Choi, and Honglak Lee.
\newblock Inferring semantic layout for hierarchical text-to-image synthesis.
\newblock In {\em Proceedings of the IEEE conference on computer vision and
  pattern recognition}, pages 7986--7994, 2018.

\bibitem{huang2022multimodal}
Xun Huang, Arun Mallya, Ting-Chun Wang, and Ming-Yu Liu.
\newblock Multimodal conditional image synthesis with product-of-experts gans.
\newblock In {\em European Conference on Computer Vision}, pages 91--109.
  Springer, 2022.

\bibitem{johnson2018image}
Justin Johnson, Agrim Gupta, and Li Fei-Fei.
\newblock Image generation from scene graphs.
\newblock In {\em Proceedings of the IEEE conference on computer vision and
  pattern recognition}, pages 1219--1228, 2018.

\bibitem{koh2021text}
Jing~Yu Koh, Jason Baldridge, Honglak Lee, and Yinfei Yang.
\newblock Text-to-image generation grounded by fine-grained user attention.
\newblock In {\em WACV}, pages 237--246, 2021.

\bibitem{li2020image}
Bowen Li, Xiaojuan Qi, Philip~HS Torr, and Thomas Lukasiewicz.
\newblock Image-to-image translation with text guidance.
\newblock {\em arXiv preprint arXiv:2002.05235}, 2020.

\bibitem{li2019object}
Wenbo Li, Pengchuan Zhang, Lei Zhang, Qiuyuan Huang, Xiaodong He, Siwei Lyu,
  and Jianfeng Gao.
\newblock Object-driven text-to-image synthesis via adversarial training.
\newblock In {\em Proceedings of the IEEE/CVF Conference on Computer Vision and
  Pattern Recognition}, pages 12174--12182, 2019.

\bibitem{li2020bachgan}
Yandong Li, Yu Cheng, Zhe Gan, Licheng Yu, Liqiang Wang, and Jingjing Liu.
\newblock Bachgan: High-resolution image synthesis from salient object layout.
\newblock In {\em Proceedings of the IEEE/CVF Conference on Computer Vision and
  Pattern Recognition}, pages 8365--8374, 2020.

\bibitem{li2021image}
Zejian Li, Jingyu Wu, Immanuel Koh, Yongchuan Tang, and Lingyun Sun.
\newblock Image synthesis from layout with locality-aware mask adaption.
\newblock In {\em Proceedings of the IEEE/CVF International Conference on
  Computer Vision}, pages 13819--13828, 2021.

\bibitem{lin2014microsoft}
Tsung-Yi Lin, Michael Maire, Serge Belongie, James Hays, Pietro Perona, Deva
  Ramanan, Piotr Doll{\'a}r, and C~Lawrence Zitnick.
\newblock Microsoft coco: Common objects in context.
\newblock In {\em ECCV}, 2014.

\bibitem{liu2022pseudo}
Luping Liu, Yi Ren, Zhijie Lin, and Zhou Zhao.
\newblock Pseudo numerical methods for diffusion models on manifolds.
\newblock {\em arXiv preprint arXiv:2202.09778}, 2022.

\bibitem{loshchilov2017decoupled}
Ilya Loshchilov and Frank Hutter.
\newblock Decoupled weight decay regularization.
\newblock {\em arXiv preprint arXiv:1711.05101}, 2017.

\bibitem{parmar2022aliased}
Gaurav Parmar, Richard Zhang, and Jun-Yan Zhu.
\newblock On aliased resizing and surprising subtleties in gan evaluation.
\newblock In {\em Proceedings of the IEEE/CVF Conference on Computer Vision and
  Pattern Recognition}, pages 11410--11420, 2022.

\bibitem{pavllo2020controlling}
Dario Pavllo, Aurelien Lucchi, and Thomas Hofmann.
\newblock Controlling style and semantics in weakly-supervised image
  generation.
\newblock In {\em European conference on computer vision}, pages 482--499.
  Springer, 2020.

\bibitem{pont2020connecting}
Jordi Pont-Tuset, Jasper Uijlings, Soravit Changpinyo, Radu Soricut, and
  Vittorio Ferrari.
\newblock Connecting vision and language with localized narratives.
\newblock In {\em European conference on computer vision}, pages 647--664.
  Springer, 2020.

\bibitem{radford2021learning}
Alec Radford, Jong~Wook Kim, Chris Hallacy, Aditya Ramesh, Gabriel Goh,
  Sandhini Agarwal, Girish Sastry, Amanda Askell, Pamela Mishkin, Jack Clark,
  et~al.
\newblock Learning transferable visual models from natural language
  supervision.
\newblock {\em arXiv preprint arXiv:2103.00020}, 2021.

\bibitem{ramesh2022hierarchical}
Aditya Ramesh, Prafulla Dhariwal, Alex Nichol, Casey Chu, and Mark Chen.
\newblock Hierarchical text-conditional image generation with clip latents.
\newblock {\em arXiv preprint arXiv:2204.06125}, 2022.

\bibitem{ramesh2021zero}
Aditya Ramesh, Mikhail Pavlov, Gabriel Goh, Scott Gray, Chelsea Voss, Alec
  Radford, Mark Chen, and Ilya Sutskever.
\newblock Zero-shot text-to-image generation.
\newblock In {\em International Conference on Machine Learning}, pages
  8821--8831. PMLR, 2021.

\bibitem{reed2016generative}
Scott Reed, Zeynep Akata, Xinchen Yan, Lajanugen Logeswaran, Bernt Schiele, and
  Honglak Lee.
\newblock Generative adversarial text to image synthesis.
\newblock In {\em International conference on machine learning}, pages
  1060--1069. PMLR, 2016.

\bibitem{rombach2022high}
Robin Rombach, Andreas Blattmann, Dominik Lorenz, Patrick Esser, and Bj{\"o}rn
  Ommer.
\newblock High-resolution image synthesis with latent diffusion models.
\newblock In {\em Proceedings of the IEEE/CVF Conference on Computer Vision and
  Pattern Recognition}, pages 10684--10695, 2022.

\bibitem{ronneberger2015u}
Olaf Ronneberger, Philipp Fischer, and Thomas Brox.
\newblock U-net: Convolutional networks for biomedical image segmentation.
\newblock In {\em International Conference on Medical image computing and
  computer-assisted intervention}, pages 234--241. Springer, 2015.

\bibitem{saharia2022photorealistic}
Chitwan Saharia, William Chan, Saurabh Saxena, Lala Li, Jay Whang, Emily
  Denton, Seyed Kamyar~Seyed Ghasemipour, Burcu~Karagol Ayan, S~Sara Mahdavi,
  Rapha~Gontijo Lopes, et~al.
\newblock Photorealistic text-to-image diffusion models with deep language
  understanding.
\newblock {\em arXiv preprint arXiv:2205.11487}, 2022.

\bibitem{schuhmann2022laion}
Christoph Schuhmann, Romain Beaumont, Cade~W Gordon, Ross Wightman, Theo
  Coombes, Aarush Katta, Clayton Mullis, Patrick Schramowski, Srivatsa~R
  Kundurthy, Katherine Crowson, et~al.
\newblock Laion-5b: An open large-scale dataset for training next generation
  image-text models.
\newblock In {\em Thirty-sixth Conference on Neural Information Processing
  Systems Datasets and Benchmarks Track}, 2022.

\bibitem{sun2019image}
Wei Sun and Tianfu Wu.
\newblock Image synthesis from reconfigurable layout and style.
\newblock In {\em Proceedings of the IEEE/CVF International Conference on
  Computer Vision}, pages 10531--10540, 2019.

\bibitem{sylvain2021object}
Tristan Sylvain, Pengchuan Zhang, Yoshua Bengio, R~Devon Hjelm, and Shikhar
  Sharma.
\newblock Object-centric image generation from layouts.
\newblock In {\em Proceedings of the AAAI Conference on Artificial
  Intelligence}, 2021.

\bibitem{wang2022git}
Jianfeng Wang, Zhengyuan Yang, Xiaowei Hu, Linjie Li, Kevin Lin, Zhe Gan,
  Zicheng Liu, Ce Liu, and Lijuan Wang.
\newblock Git: A generative image-to-text transformer for vision and language.
\newblock {\em arXiv preprint arXiv:2205.14100}, 2022.

\bibitem{wang2022ofa}
Peng Wang, An Yang, Rui Men, Junyang Lin, Shuai Bai, Zhikang Li, Jianxin Ma,
  Chang Zhou, Jingren Zhou, and Hongxia Yang.
\newblock Ofa: Unifying architectures, tasks, and modalities through a simple
  sequence-to-sequence learning framework.
\newblock In {\em International Conference on Machine Learning}, pages
  23318--23340. PMLR, 2022.

\bibitem{xu2018attngan}
Tao Xu, Pengchuan Zhang, Qiuyuan Huang, Han Zhang, Zhe Gan, Xiaolei Huang, and
  Xiaodong He.
\newblock Attngan: Fine-grained text to image generation with attentional
  generative adversarial networks.
\newblock In {\em Proceedings of the IEEE conference on computer vision and
  pattern recognition}, pages 1316--1324, 2018.

\bibitem{yang2022unitab}
Zhengyuan Yang, Zhe Gan, Jianfeng Wang, Xiaowei Hu, Faisal Ahmed, Zicheng Liu,
  Yumao Lu, and Lijuan Wang.
\newblock Unitab: Unifying text and box outputs for grounded vision-language
  modeling.
\newblock In {\em European Conference on Computer Vision}, pages 521--539.
  Springer, 2022.

\bibitem{yang2022modeling}
Zuopeng Yang, Daqing Liu, Chaoyue Wang, Jie Yang, and Dacheng Tao.
\newblock Modeling image composition for complex scene generation.
\newblock In {\em Proceedings of the IEEE/CVF Conference on Computer Vision and
  Pattern Recognition}, pages 7764--7773, 2022.

\bibitem{yu2022scaling}
Jiahui Yu, Yuanzhong Xu, Jing~Yu Koh, Thang Luong, Gunjan Baid, Zirui Wang,
  Vijay Vasudevan, Alexander Ku, Yinfei Yang, Burcu~Karagol Ayan, et~al.
\newblock Scaling autoregressive models for content-rich text-to-image
  generation.
\newblock {\em arXiv preprint arXiv:2206.10789}, 2022.

\bibitem{zhang2021cross}
Han Zhang, Jing~Yu Koh, Jason Baldridge, Honglak Lee, and Yinfei Yang.
\newblock Cross-modal contrastive learning for text-to-image generation.
\newblock In {\em Proceedings of the IEEE/CVF conference on computer vision and
  pattern recognition}, pages 833--842, 2021.

\bibitem{zhang2017stackgan}
Han Zhang, Tao Xu, Hongsheng Li, Shaoting Zhang, Xiaogang Wang, Xiaolei Huang,
  and Dimitris~N Metaxas.
\newblock Stackgan: Text to photo-realistic image synthesis with stacked
  generative adversarial networks.
\newblock In {\em Proceedings of the IEEE international conference on computer
  vision}, pages 5907--5915, 2017.

\bibitem{zhang2018stackgan++}
Han Zhang, Tao Xu, Hongsheng Li, Shaoting Zhang, Xiaogang Wang, Xiaolei Huang,
  and Dimitris~N Metaxas.
\newblock Stackgan++: Realistic image synthesis with stacked generative
  adversarial networks.
\newblock {\em IEEE transactions on pattern analysis and machine intelligence},
  41(8):1947--1962, 2018.

\bibitem{zhao2019image}
Bo Zhao, Lili Meng, Weidong Yin, and Leonid Sigal.
\newblock Image generation from layout.
\newblock In {\em Proceedings of the IEEE/CVF Conference on Computer Vision and
  Pattern Recognition}, pages 8584--8593, 2019.

\bibitem{zhou2022detecting}
Xingyi Zhou, Rohit Girdhar, Armand Joulin, Philipp Kr{\"a}henb{\"u}hl, and
  Ishan Misra.
\newblock Detecting twenty-thousand classes using image-level supervision.
\newblock In {\em European Conference on Computer Vision}, pages 350--368.
  Springer, 2022.

\bibitem{zhou2022towards}
Yufan Zhou, Ruiyi Zhang, Changyou Chen, Chunyuan Li, Chris Tensmeyer, Tong Yu,
  Jiuxiang Gu, Jinhui Xu, and Tong Sun.
\newblock Towards language-free training for text-to-image generation.
\newblock In {\em Proceedings of the IEEE/CVF Conference on Computer Vision and
  Pattern Recognition}, pages 17907--17917, 2022.

\end{thebibliography}
}

\end{document}